\documentclass[conference]{IEEEtran}
\IEEEoverridecommandlockouts
\usepackage{cite}
\usepackage{amsmath,amssymb,amsfonts}
\usepackage{algorithmic}
\usepackage{graphicx}
\usepackage{textcomp}
\usepackage{xcolor}
\usepackage{multirow}
\usepackage{booktabs}
\usepackage{url}
\usepackage{tcolorbox}
\usepackage{subfig} 
\usepackage{caption}
\usepackage{balance}
\usepackage{enumitem}
\usepackage{graphicx}
\usepackage{tabularx}

\def\BibTeX{{\rm B\kern-.05em{\sc i\kern-.025em b}\kern-.08em
		T\kern-.1667em\lower.7ex\hbox{E}\kern-.125emX}}

\usepackage{makecell}
\usepackage{threeparttable}
\makeatletter  
\newif\if@restonecol  
\renewcommand\footnoterule{%
	\kern-3\p@
	\hrule\@width\columnwidth
	\kern2.6\p@}

\usepackage{colortbl}
\definecolor{mygray}{gray}{.9}
\definecolor{myother}{gray}{0.7}

\usepackage{url}

\DeclareRobustCommand*{\IEEEauthorrefmark}[1]{%
	\raisebox{0pt}[0pt][0pt]{\textsuperscript{\footnotesize\ensuremath{#1}}}}

\begin{document}
	
	
	
	
	
	\title{An Empirical Study towards Characterizing Deep Learning Development and Deployment across Different Frameworks and Platforms}
	
	\author{\IEEEauthorblockN{
			Qianyu Guo\IEEEauthorrefmark{1},
			Sen Chen\IEEEauthorrefmark{2}$^*$\thanks{$^*$Sen Chen (chensen@ntu.edu.sg) and XiaohongLi (xiaohongli@tju.edu.cn) are the corresponding authors.},
			Xiaofei Xie\IEEEauthorrefmark{2},
			Lei Ma\IEEEauthorrefmark{3},
			Qiang Hu\IEEEauthorrefmark{3},
			Hongtao Liu\IEEEauthorrefmark{1},\\
			Yang Liu\IEEEauthorrefmark{2},
			Jianjun Zhao\IEEEauthorrefmark{3},
			Xiaohong Li\IEEEauthorrefmark{1}$^*$
		}
		\IEEEauthorblockA{\IEEEauthorrefmark{1}College of Intelligence and Computing, Tianjin University, China \\
			\IEEEauthorrefmark{2}Nanyang Technological University, Singapore
			\IEEEauthorrefmark{3}Kyushu University, Japan
	}}
	
	\maketitle
	\begin{abstract}
		Deep Learning (DL) has recently achieved tremendous success. A variety of DL frameworks and platforms play a key role to catalyze such progress. However, the differences in architecture designs and implementations of existing frameworks and platforms bring new challenges for DL software development and deployment. Till now, there is no study on how various mainstream frameworks and platforms influence both DL software development and deployment in practice. 
		
		To fill this gap, we take the first step towards understanding how the most widely-used DL frameworks and platforms support the DL software development and deployment. We conduct a systematic study on these frameworks and platforms by using two types of DNN architectures and three popular datasets. (1) For development process, we investigate the prediction accuracy under the same runtime training configuration or same model weights/biases.
		We also study the adversarial robustness of trained models by leveraging the existing adversarial attack techniques. The experimental results show that the {computing differences} across frameworks could result in an obvious prediction accuracy decline, which should draw the attention of DL developers. (2) For deployment process, we investigate the prediction accuracy and performance (refers to time cost and memory consumption) when the trained models are migrated/quantized from PC to real mobile devices and web browsers.
		The DL platform study unveils that the migration and quantization still suffer from compatibility and reliability issues. Meanwhile, we find several DL software bugs by using the results as a benchmark. We further validate the results through bug confirmation from stakeholders and industrial positive feedback to highlight the implications of our study. Through our study, we summarize practical guidelines, identify challenges and pinpoint new research directions, such as understanding the characteristics of DL frameworks and platforms, avoiding compatibility and reliability issues, detecting DL software bugs, and reducing time cost and memory consumption towards developing and deploying high quality DL systems effectively.
	\end{abstract}
	
	
	
	\section{Introduction}\label{sec:intro}
	With the big data explosion and hardware evolution over the past decade, deep learning (DL) has achieved tremendous success in many cutting-edge domains, such as real-time strategy game~\cite{dota2}, image processing~\cite{resnet}, speech and language processing~\cite{jurafsky2014speech}, and autonomous vehicle~\cite{chen2015deepdriving}.
	The deep neural network (DNN)~\cite{dnn} plays a key role behind such recent success of DL applications.
	It automatically learns the decision logic from the training data, which is represented in the form of a neural network and the connection strengths among neurons.
	
	To transfer the learning theory into practice, a number of DL frameworks (e.g., \textsc{\small TensorFlow}~\cite{tensorflow} and \textsc{\small PyTorch}~\cite{paszke2017automatic}) are developed towards realizing the demands of intelligent software. 
	Although most of the existing DL frameworks share either static or dynamic computation paradigms~\cite{gupta2004static}, the detailed architecture design and implementation of frameworks are quite different. 
	Actually, even the same DNN architecture design with exactly the same runtime configuration {(i.e., random seed for initialization and hyper parameters for training)} might result in different decisions when implemented under different DL frameworks, which brings new challenges for DL software development process.
	Several DL benchmarking studies have focused on some basic metrics of DL frameworks~\cite{bahrampour2015comparative,awan2017depth,coleman2017dawnbench,shatnawi2018comparative}, such as training and testing accuracy, the influence of hardwares (i.e., GPU and CPU), and also compared {different frameworks with their default configuration settings and training data specific parameters~\cite{liu2018benchmarking}.}
	However, there lacks an empirical study on the impacts that various DL frameworks under the same runtime configuration or same model weigths/biases have on the DL software development process.
	
	Moreover, with the great demand on deploying the DL software to different platforms, it further poses new challenges when DL models on the PC platform are migrated, quantized, and deployed on other platforms such as real mobile devices and web browsers. 
	While a computational intensive DL software could be executed efficiently on PC platform with the GPU support, such DL models usually cannot be directly deployed and executed on other platforms supported by small devices {due to various limitations, such as the computation power, memory size and energy.}
	Therefore, some DL frameworks are specifically designed for mobile platforms, such as \textsc{\small TensorFlow Lite}~\cite{tensorlite} for Android and  \textsc{\small Core ML}~\cite{coreml} for iOS. 
	Similarly, \textsc{\small TensorFlow.js}~\cite{tensorjs} for web DL applications is also proposed.
	Meanwhile, in terms of mobile devices, it is a common practice that a DL model needs to undergo a quantization process before the deployment, considering the limited resources of memory and energy on mobile devices~\cite{quantization}.
	There lacks an empirical study focusing on the process of migration and quantization on mobile and web platforms.
	
	Although the diverse DL frameworks and platforms promote the evolution of DL software, understanding the characteristics of them becomes a time-consuming task for DL software developers and researchers. 
	Moreover, the differences compared with the traditional software brings new challenges for DL software development and deployment processes.
	These challenges include that 
	(1) for the development process, there lacks a deep understanding of various frameworks under a) the training and prediction accuracy given the same runtime configuration; b) the prediction accuracy given the same model weights/biases; and c) the robustness of trained models.
	(2) For the deployment process, when deploying the trained models from PC/Server to different platforms, there lacks a benchmarking understanding of the migration and quantization processes, such as the impacts on prediction accuracy, performance (i.e., time cost and memory consumption).
	
	To address the aforementioned challenges, with an over ten man-month effort, we design and perform an empirical study on the state-of-the-art DL frameworks and platforms from two aspects to investigate the following research questions.
	
	\vspace{1mm}
	\noindent{(1) As for the development process:}
	\begin{itemize}[noitemsep,topsep=0pt,leftmargin=*]
		\item {\textbf{RQ1}}: 
		\textbf{Accuracy on different frameworks.}
		Given the same runtime configuration or same model weights/biases,
		what are the differences of training and prediction accuracy when implemented with different DL frameworks? 
		
		\item {\textbf{RQ2}}: 
		\textbf{Adversarial robustness of trained models.}
		{Do DL models trained from different DL frameworks exhibit the same adversarial robustness against adversarial examples?}
	\end{itemize}
	\vspace{1mm}
	\noindent{(2) As for the deployment process:}
	\begin{itemize}[noitemsep,topsep=0pt,leftmargin=*]
		\item {\textbf{RQ3}}: 
		\textbf{Performance after migration and quantization.}
		What are the differences of performance (i.e., time cost and memory consumption) in the capabilities of supporting DL software when migrating or quantizing the trained models to the real mobile devices and web browsers?
		
		\item {\textbf{RQ4}}: 
		\textbf{Prediction accuracy after migration and quantization.} 
		{Given the same trained DL model, what is the prediction accuracy of the migrated model for mobile and web platforms? 
			How do quantization methods influence the prediction accuracy of quantized model on mobile devices?}
	\end{itemize}
	
	\vspace{1mm}
	Through answering these research questions, we aim to characterize the impacts of current DL frameworks and platforms on DL software development and deployment processes, and provide practical guidelines to developers and researchers from different research communities such as SE and AI fields and under different practical scenarios. 
	
	In summary, we make the following main contributions:
	
	\begin{itemize}[noitemsep,topsep=0pt,leftmargin=*]
		\item To the best of our knowledge, this is the first empirical study on how the current DL frameworks and platforms influence the development and deployment processes, especially for the study on the migration and quantization processes on different DL platforms.
		
		\item For the development process, we find the computing differences across frameworks might result in an obvious prediction accuracy decline. That would be a great warning to the DL developers and SE testing researchers.
		Our further investigation finds the adverarial robustness of trained models from different frameworks is also different.
		
		\item For the deployment process, 
		6 real mobile devices and 3 web browsers have different performance in capabilities of supporting DL software.
		Mobile platforms have a better prediction accuracy of migration than that of current web platforms, and the web platforms have an obvious compatibility issue (i.e., prediction accuracy drops over 5\%). We find a real bug according to the phenomenon and report to the stakeholders. It is confirmed and appreciated by developers. More bug information can be found on our website~\cite{our}.
		Moreover, the quantization of mobile platforms suffer from significant reliability issues on our generated testing dataset, and it is hard to trigger such issue by the widely-used original testing data. 
		That would motivate the SE researchers to conduct a further test in this field.
		
		\item We also conduct an online questionnaire~\cite{survey} to validate the usefulness of our study, and receive 20 industrial positive feedback from the AI research teams in Baidu China, Huawei Signapore, and NetEase China, which confirms the usefulness of our study.
		In addition, we make all generated testing dataset used in our evaluation on migrated and quantized models publicly available~\cite{our}, to facilitate further study towards more systematic investigation.
		
		\item 
		We highlight the challenges and propose new research directions.
		Meanwhile, our empirical study can be used as a benchmark and baseline for issues and bugs detection to evaluate new DL frameworks and platforms.
	\end{itemize}

	\section{Background}
	In this section, we briefly introduce the current practice of DL software development and deployment.
	
	\subsection{DL Software Development}
	DL software development contains several phases (e.g., data collection and labelling, DNN design, runtime training, and testing/validation). DL developers design the DNN architecture and specify runtime configuration (e.g., random seed and hyper parameters) before training on selected dataset. It is a common practice that using the state-of-the-art DL frameworks to accomplish training, followed by the validation/testing stage for accuracy evaluation on the trained models.
	
	\begin{figure*}[t]
		\centering
		\includegraphics[width=0.98\textwidth]{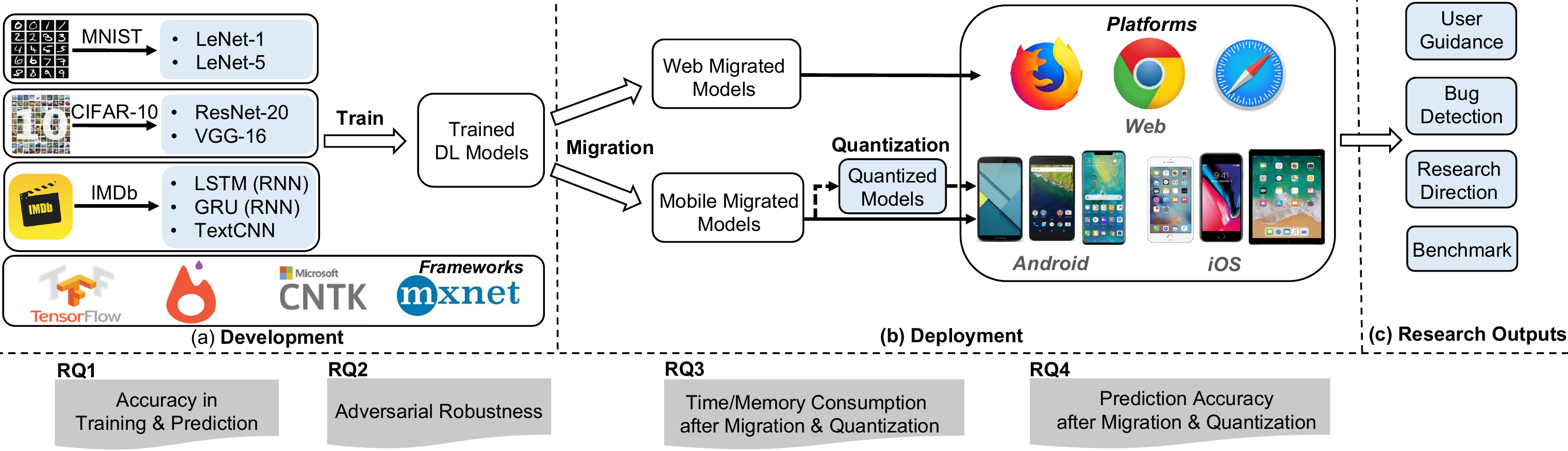}
		\caption{Overview of our study}
		\label{fig:overview}
	\end{figure*}
	
	\subsection{DL Software Deployment} 
	A DL software, that has been well tested and validated and reaches a certain level of quality standard, is ready to be deployed for application (e.g., web and mobile platforms). Developers need to consider calibration (e.g., migration and quantization) when deploying DL software across different platforms.
	
	For web platform, several solutions~(e.g., \textsc{\small TensorFlow.js}) are proposed for deploying DL models under the web environment. For mobile platform, although the rapid advances in system-on-chip~(SoC)~\cite{snapdragon}\cite{kirin-970}\cite{samsung-exynos-9} facilitate the AI applications for mobile use, existing trained DL models on PC could still not be directly deployed on mobile devices, due to the limitations such as computing power, memory size and energy capacity. Some lightweight solutions~(e.g.,\textsc{\small TensorFlow Lite} and \textsc{\small Core ML}) are proposed to support this migration. Moreover, it is a common practice to conduct a quantization process before deploying DL models on mobile devices, so as to reduce memory cost and computing overhead~\cite{quantization}.
	
	\textsc{\small TensorFlow} provides two options for quantization (i.e., {post-training quantization}~\cite{post-training-quantization} and {quantization aware training}~\cite{quantization-aware-training}), both of which fixedly convert model weights to 8-bits integers from floating points, using a linear weights representation. \textsc{\small Core ML} supports flexible quantization modes~\cite{coreml-quantization} (i.e., linear, linear\_lut, kmeans\_lut, and custom\_lut), along with a \textit{nbits} option, which allows to customize the bits of per quantized weight~(e.g., 32-bits to 16/8/4-bits).
	
	\section{Overview}
	In this section, we briefly introduce the overview of our study and the evaluation objects and metrics.
	\subsection{Study Design}\label{sec:setup}

	Fig.~\ref{fig:overview} shows the overview of our study, which contains two main phases (i.e., development and deployment) to answer the four research questions. For the development process, we investigate the training and prediction accuracy and adversarial robustness of trained models across different frameworks. To achieve these goals, we select 4 widely-used frameworks (i.e., \textsc{\small TensorFlow}~\cite{tensorflow}, \textsc{\small PyTorch}~\cite{paszke2017automatic}, \textsc{\small CNTK}~\cite{seide2016cntk}, and \textsc{\small MXNet}~\cite{mxnet}) as our evaluation objects, and use 3 publicly available datasets (i.e., MNIST, CIFAR-10, and IMDb) for training and prediction on each of them.
	Correspondingly, we choose 7 popular DNN models (i.e., LeNet-1, LeNet-5~\cite{lenet-fam}, RestNet-20~\cite{resnet}, VGG-16~\cite{vgg}, TextCNN~\cite{textcnn}, LSTM (RNN)~\cite{lstm} and GRU (RNN)~\cite{gru}) for inspection, including CNN and RNN architectures. 
	
	
	For the deployment process, we focus on the model performance and prediction accuracy after migrated and quantized to different platforms.
	To conduct these evaluations, 2 popular platforms are selected to evaluate (1) 3 popular web browsers (Chrome, Firefox, and Safari); and (2) 6 real mobile devices: 3 Android devices~(i.e., Nexus 6, Nexus 6P, and HUAWEI Mate 20X) and 3 iOS devices~(i.e., iPhone 6S, iPhone 8, and iPad Pro). We migrate and deploy the models trained in the development process to the two types of platforms. Meanwhile, we follow the common practice to further conduct model quantization for mobile devices to investigate their performance and prediction accuracy.
	
	\subsection{DL Frameworks and Platforms}\label{sec:objects}
	DL frameworks play an important role to bridge the DL theory to the practice of DL software. We select the most updated versions of four representative frameworks (i.e., \textsc{\small TensorFlow}-1.12.0 from Google, \textsc{\small PyTorch}-0.4.1 from Facebook, \textsc{\small CNTK}-2.6 from Microsoft, and \textsc{\small MXNet}-1.4.0 maintained by Apache) for investigation, where \textsc{\small TensorFlow} and \textsc{\small CNTK} adopt the static computational graph paradigm, while \textsc{\small PyTorch} follows a dynamic computational paradigm. \textsc{\small MXNet} adopts both two types.
	We investigate three DL platforms, where an urgent demand on DL software solutions exists from industry. (1) PC, the mainstream platform where most DL models are trained. (2) Mobile platform such as Android and iOS mobile devices. (3) Web platform~(i.e., Chrome, Firefox, and Safari).

	\subsection{Datasets and DNN Models}\label{sec:dataset:model}
	In order to conduct our study, we select three publicly available datasets (i.e., MNIST~\cite{mnist}, CIFAR-10~\cite{cifar}, and IMDb~\cite{imdb}) for training and prediction, all of them are widely used in DL community. 
	{For each dataset, we follow the best DL practice and choose diverse DNN models (i.e., LeNet-1, LeNet-5, RestNet-20, VGG-16, TextCNN, LSTM (RNN) and GRU (RNN)) that are able to achieve competitive results in terms of training and testing accuracy.} 
	We detail the hyper-parameters of each DNN model on specific dataset on ~\cite{our}.
	
	MNIST is a collection of gray-scale images used for hand-written digit recognition.
	For MNIST, we use two well-known models from the LeNet family (i.e., LeNet-1 and LeNet-5~\cite{lenet-fam}).
	CIFAR-10 is a collection of colored images~(e.g., airplane, automobile, and bird) for object classification. 
	For CIFAR-10, we select two popular DNN models (i.e., ResNet-20~\cite{resnet} and VGG-16~\cite{vgg}) for inspection, both of which could achieve competitive prediction accuracy.
	IMDb is a collection of text-based movie reviews from the online database IMDb~\cite{imdb}, which is widely used for text sentiment classification in the field of natural language processing.
	As for IMDb, we select a CNN-based model TextCNN~\cite{textcnn} and an RNN-based model TextRNN for inspection, both of which are classical models in NLP.
	There are two types of implementations for TextRNN (i.e., LSTM~\cite{greff2017lstm} and GRU~\cite{chung2014empirical}). 
	
	\subsection{Evaluation Metrics}\label{sec:metrics}
	
	
	\noindent{\bf Accuracy in Training and Prediction.} 
	At the training stage, we first ensure the same runtime configuration across different frameworks.
	Then we train the models with multiple combinations of hyper parameters on these frameworks, and evaluate the training and validation accuracy in this stage.
	We select one combination as example shown in this paper, which achieves comparable training accuracy for all selected frameworks.
	For prediction stage, we evaluate the accuracy and time cost using the testing data. 
	{Particularly, to investigate the computing difference of different frameworks, we further ensure the same weights/biases of the same model by using \textsc{\small MMdnn}~\cite{mmdnn} across different frameworks, and evaluate the prediction accuracy.}

	

	
	\noindent{\bf Adversarial Robustness.}
	Robustness indicates the quality and security of the trained DL models~\cite{katz2017reluplex,ma2018combinatorial}. We focus on a typical robustness property \emph{adversarial robustness} in this paper. The adversarial robustness concerns whether there exists an example  ${x}'$~(close to a given input $x$) that ${x}$ and ${x}'$ are misclassified by the DNN. Such ${x}'$, once exists, is called an \emph{adversarial example} of ${x}$ and the DNN is not adversarial robust at ${x}$. Formally, a DNN's adversarial robustness could be analyzed by checking the $d$-local-robustness at an input ${x}$ w.r.t a distance parameter $d$ if we have the following relation~\cite{katz2017reluplex}:
	{$$\forall {x}':||{x}'-{x}||\leq d \Rightarrow \mathcal{C}({x})=\mathcal{C}({x}'),$$}
	where $x$ could be correctly predicted by the DNN. We follow the currently best practice in machine learning~\cite{carlini2017towards} to generate adversarial examples by exerting adversarial attacks~\cite{goodfellow6572explaining}\cite{papernot2016practical}\cite{brendel2017decision} on DL models.
	
	
	\noindent{\bf Accuracy and Performance in Migration and Quantization.}
	It is common that a DL model with complex structure could achieve competitive results on PC or cloud, but inevitably introduce large computing and memory overheads at the same time.
	When DL models are migrated from PC to web and mobile platforms, we observe the accuracy and performance~(i.e., time cost and memory consumption) change in this process.
	Moreover, to deploy such DL models on the resource-limited mobile devices, quantization is a common practice to ensure the smooth running~\cite{quantization}. We study how quantization technique influences the accuracy and time cost in prediction.
	
	\section{Empirical Study}\label{sec:study}
	In this section, we first briefly introduce the experimental environment for our study, and then we detail the numerous experiments to answer the 4 research questions highlighted in Section~\ref{sec:intro}.
	
	
	(1) For the development study, we train 7 DL models on 3 types datasets using 4 DL frameworks, respectively.
	We use multiple combinations of hyper parameters for each model in the training stage, aiming to obtain a relatively good training accuracy on each framework and avoiding over-fitting/under-fitting as much as possible. Meanwhile, we repeat each model training and testing processes 5 times. 
	(2) For the deployment study, 7 trained models from \textsc{\small TensorFlow} are migrated and executed on 3 web browsers, and 4 of them are also converted to mobile devices. 6 real mobile devices including Android/iOS devices are selected to run the 4 migrated/quantized models.
	For each web browser/mobile device, we conduct 5 parallel evaluations on each model to minimize the random impacts as much as possible.
	The study takes 10 months, including the substantial effort on model training, migration/quantization, and cross-platform evaluations.
	
	\noindent{\bf Experimental Environment.}
	We run all the PC application experiments on a high performance computer cluster.
	Each cluster node runs a GNU/Linux system with Linux kernel 4.4.0 on 2 18-core 2.3GHz Intel Xeon CPU E5-2699 with 190 GB RAM equipped with a NVIDIA Tesla P40 GPUs. Web application experiments are conducted on a laptop with 64-bit Chrome 71.0.3578.98, Firefox 64.0.2 and Safari 12.0.2.
	{The host laptop is MacBook Pro with macOS 10.14.2 on a 2.7GHz Intel Core i7 CPU with 16GB RAM.}
	The mobile application experiments are conducted on real Android devices (i.e., HUAWEI Mate 20X, HUAWEI Nexus 6P, and Motorola Nexus 6) with Android {9.0 API 28}, 7.1.1 API 25 and 8.1.0 API 27 and iOS devices (i.e., iPhone 8, iPhone 6S, and iPad Pro) with iOS 12.1.2.
	
	\subsection{RQ1: Accuracy on Different Frameworks}

	\subsubsection{Training Accuracy}\label{sec:training-performance}
	To investigate the training accuracy across different DL frameworks, 7 DNN models (i.e., LeNet-1 and LeNet-5 for MNIST, ResNet-20 and VGG-16 for CIFAR-10, TextCNN, LSTM (RNN), and GRU (RNN) for IMDb) are trained on four different frameworks. 
	For each model, we ensure the same runtime configuration on different frameworks.
	For example, we set identical learning rate (i.e., 0.05), training epochs (i.e., 200), optimizer (i.e., SGD), batch size (i.e., 128), etc. for LeNet-1 on all frameworks. Each DNN model is repeatedly trained for 5 times under each framework, and one with the highest validation accuracy is selected for comparison.
	We only demonstrate the accuracy of training and prediction by using 3 DNN models (i.e., LeNet-5, VGG-16, and GRU (RNN)) based on three data types due to the space limitation. More training plots can be found on our website~\cite{our}.

	\begin{figure}
		\centering
		\subfloat[LeNet-5]{\includegraphics[width=2.857cm,height=2.175cm]{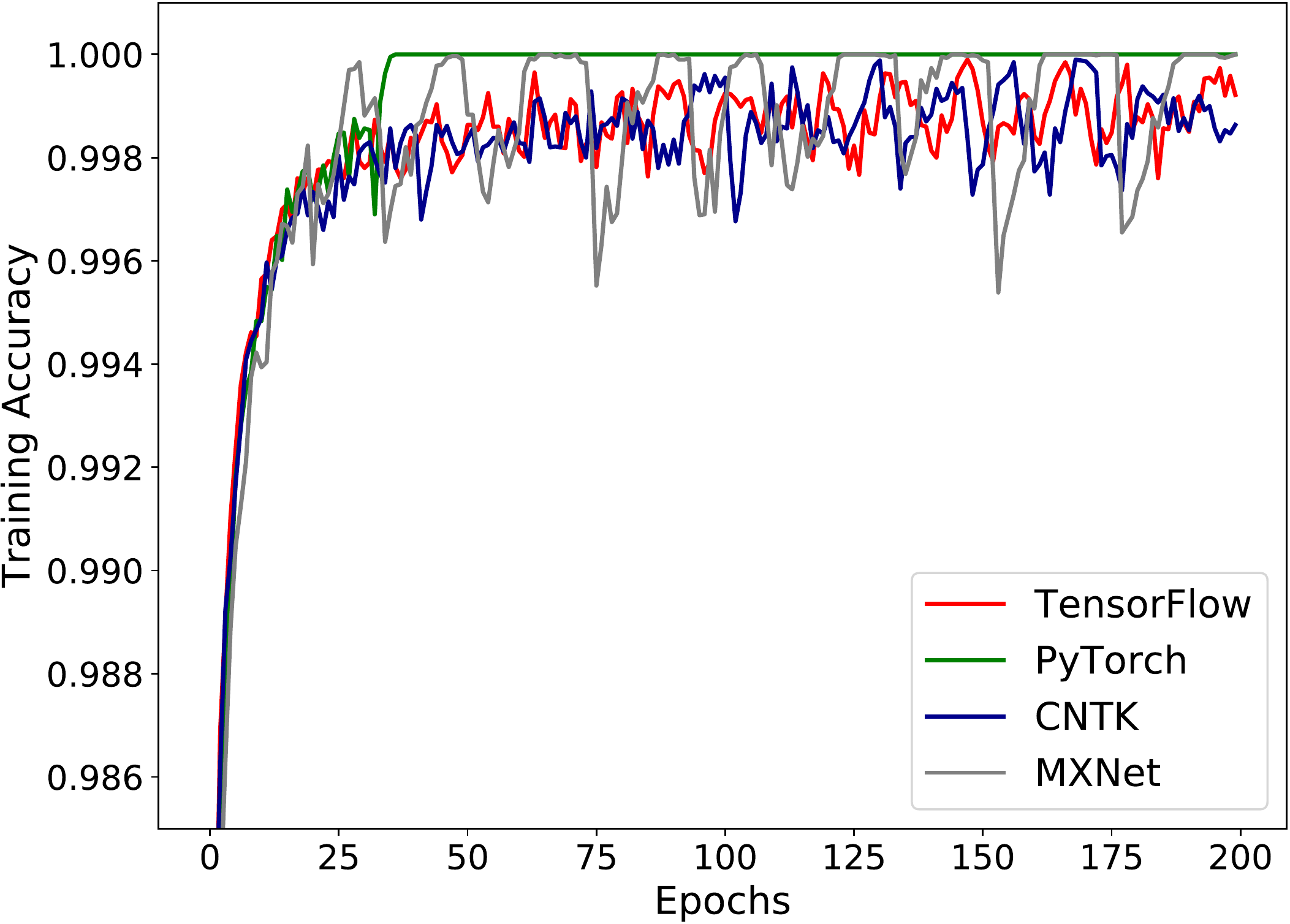}\label{fig:trainacc:fig1}}
		\subfloat[VGG-16]{\includegraphics[width=2.857cm,height=2.175cm]{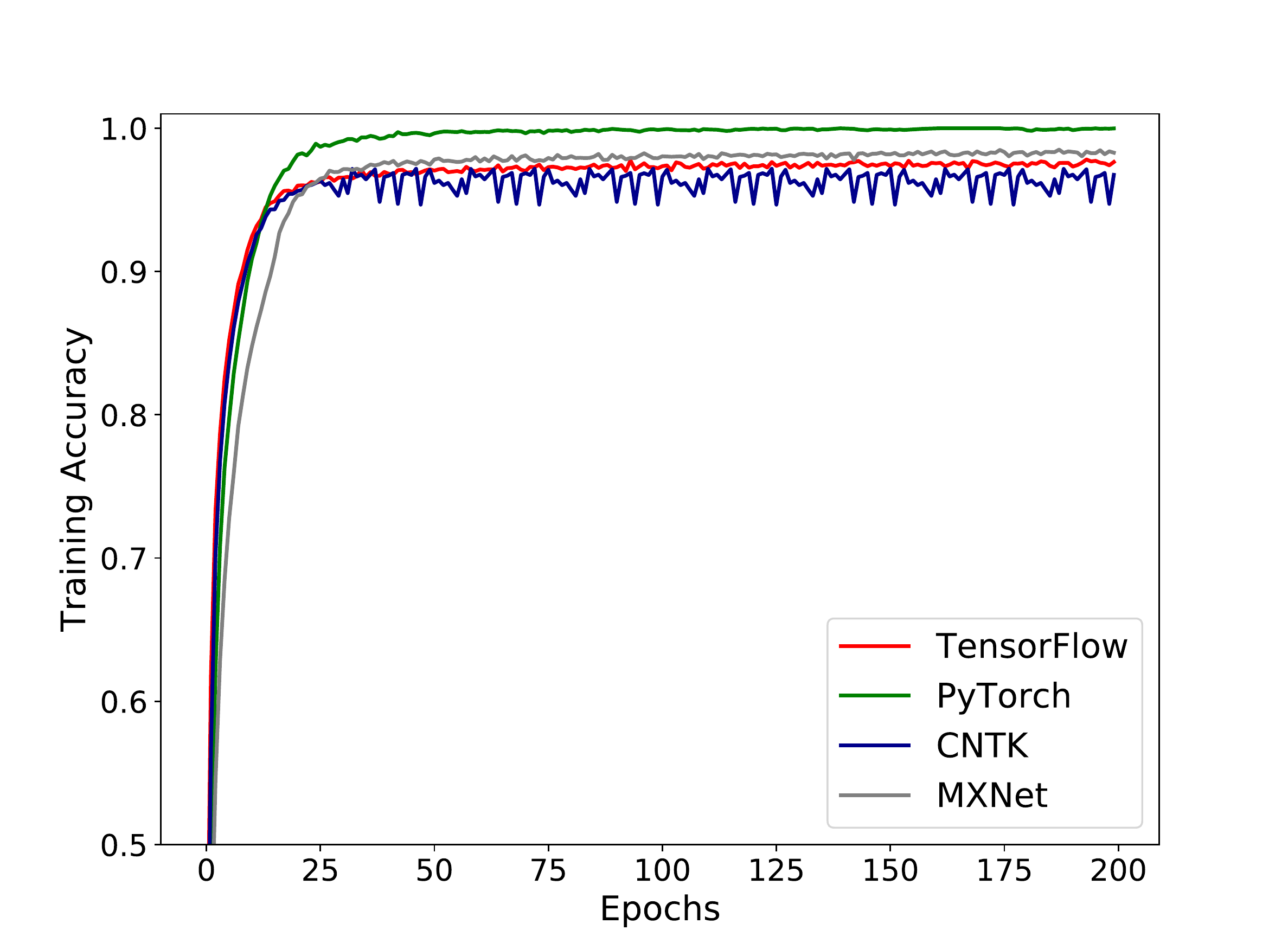}\label{fig:trainacc:fig2}}
		\subfloat[GRU]{\includegraphics[width=2.857cm,height=2.175cm]{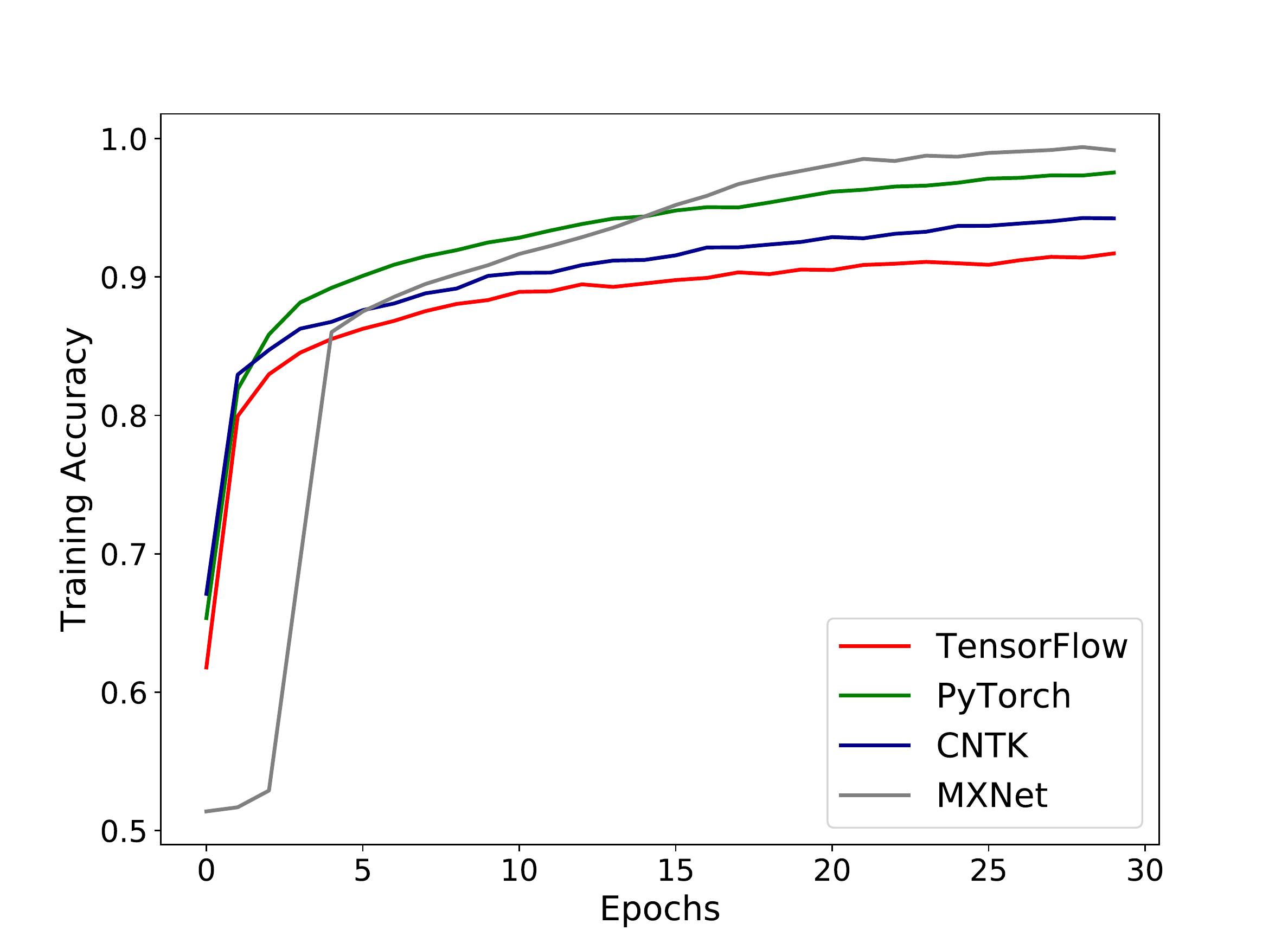}\label{fig:trainacc:fig3}}
		\caption{Training accuracy of LeNet-5, VGG-16, and GRU with different DL frameworks}
		\label{fig:trainacc}
	\end{figure}
	\begin{figure}
		\centering
		\subfloat[LeNet-5]{\includegraphics[width=2.857cm,height=2.175cm]{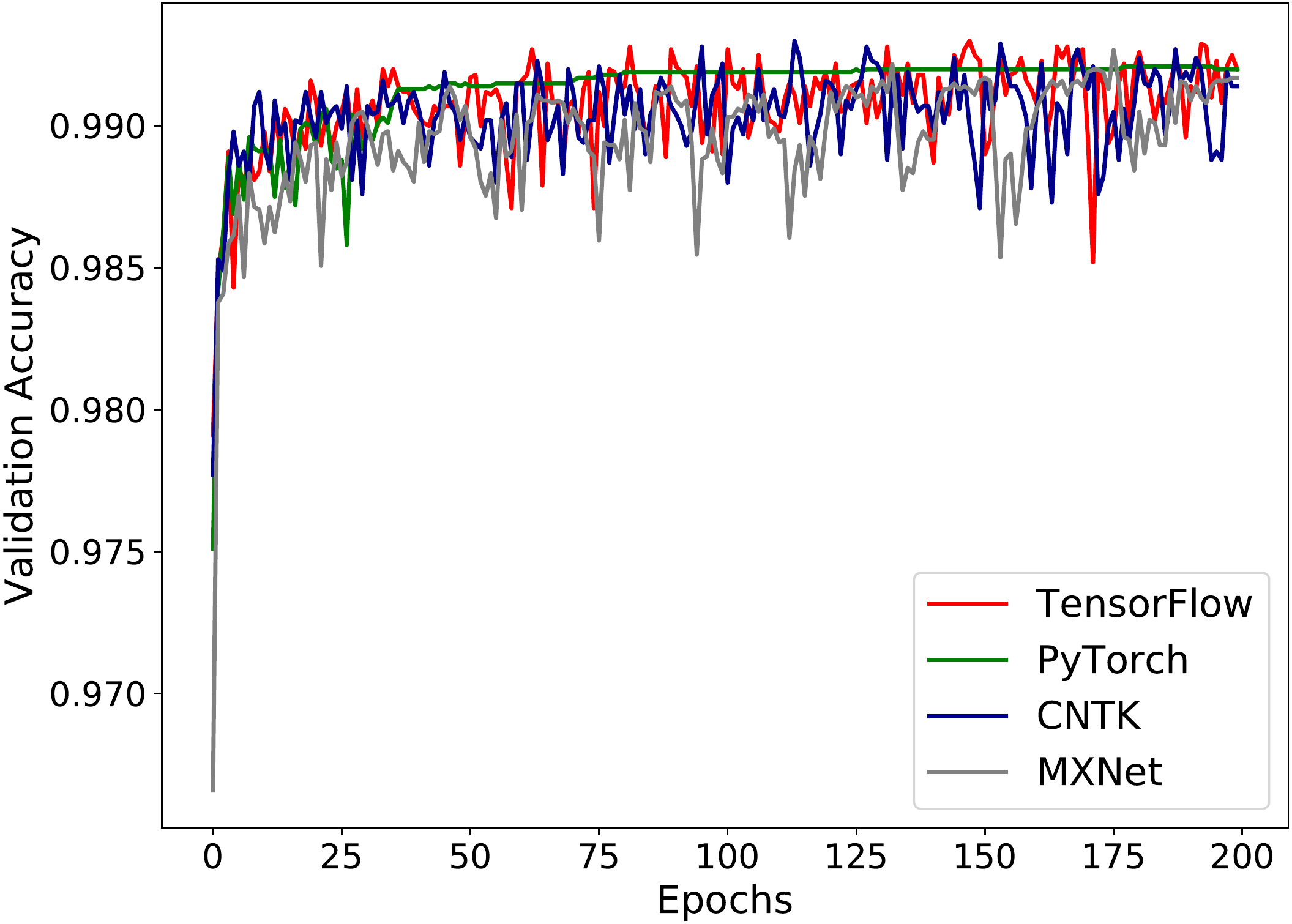}\label{fig:valacc:fig1}}
		\subfloat[VGG-16]{\includegraphics[width=2.857cm,height=2.175cm]{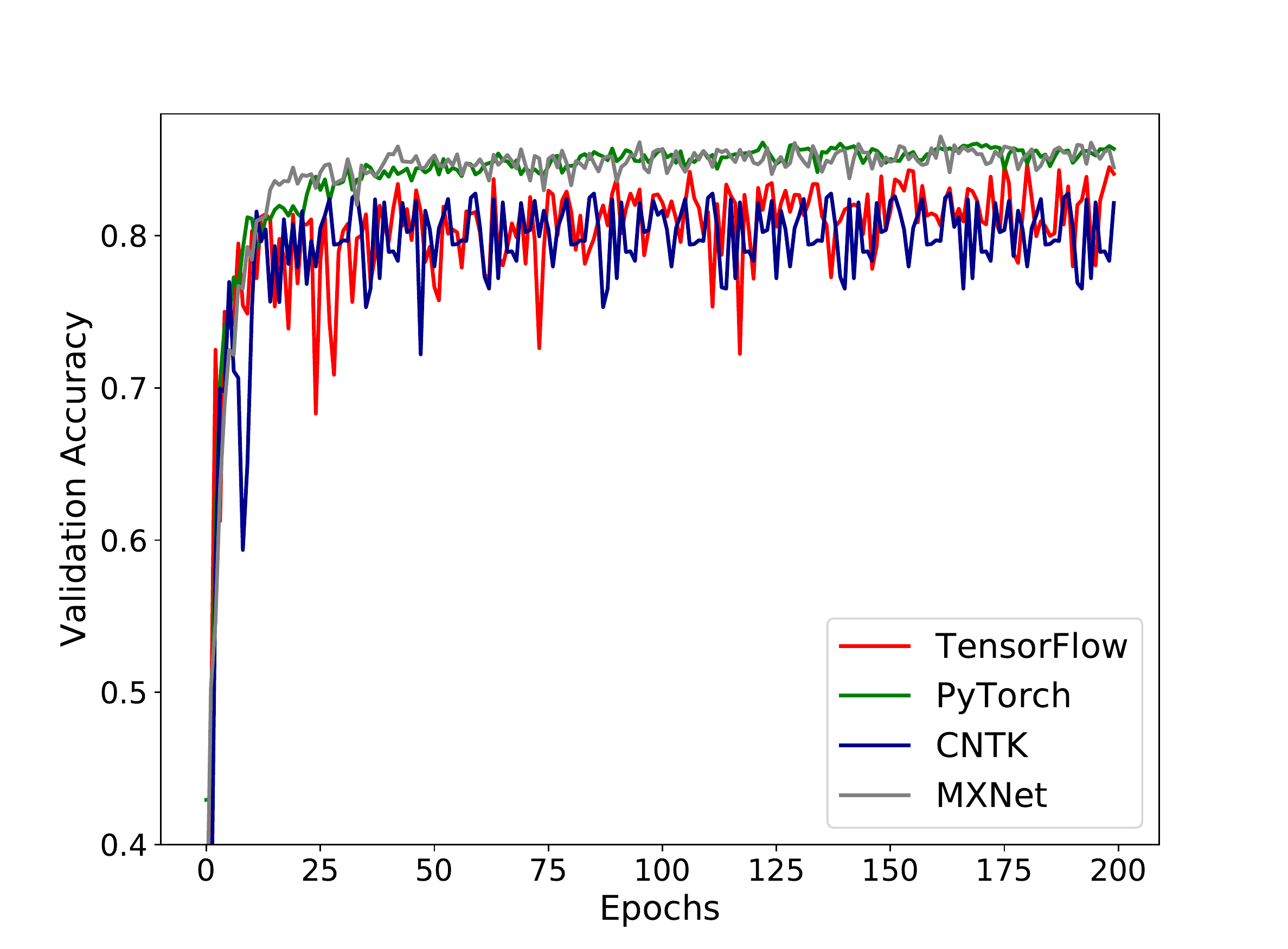}\label{fig:valacc:fig2}}
		\subfloat[GRU]{\includegraphics[width=2.857cm,height=2.175cm]{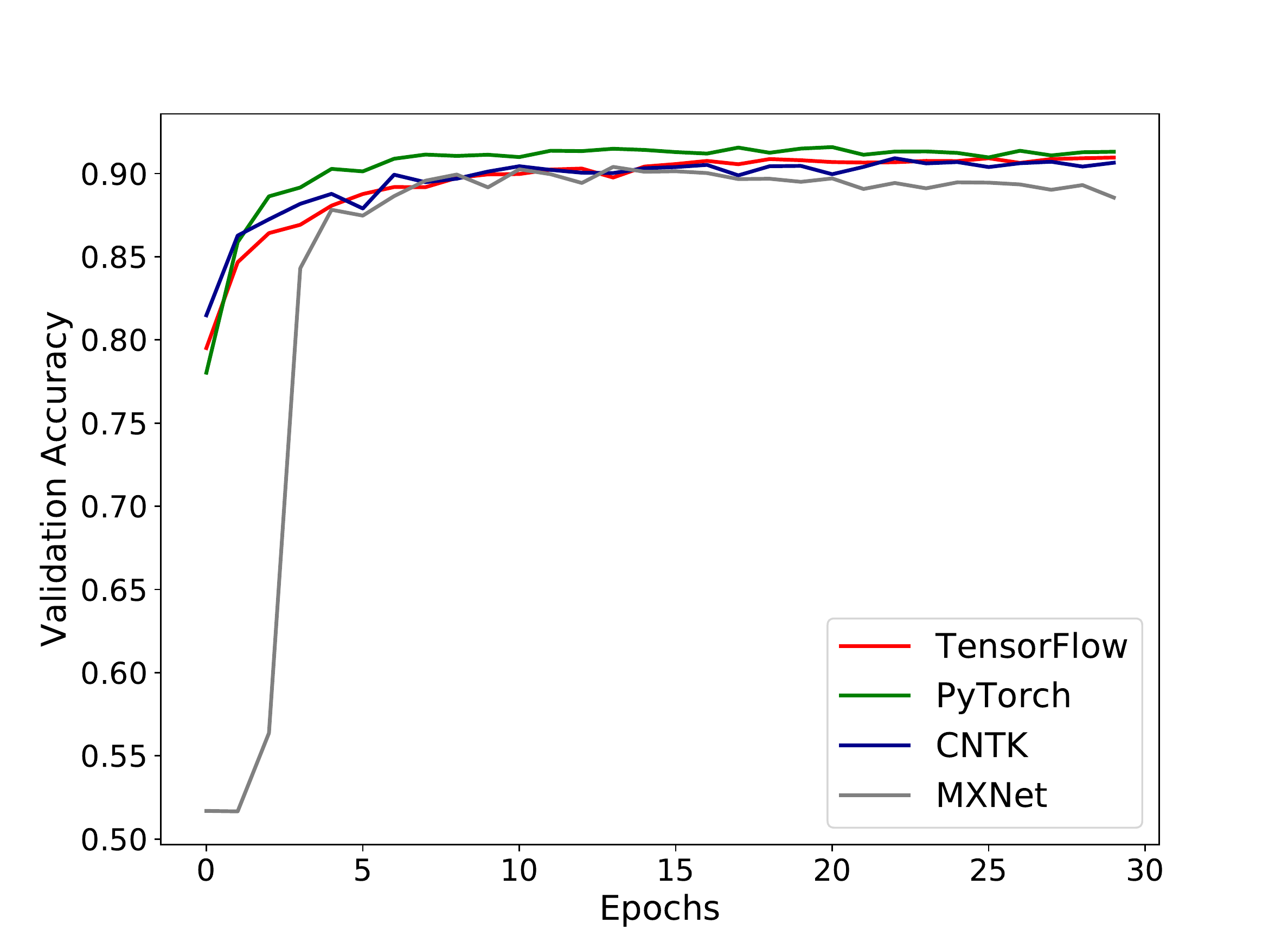}\label{fig:valacc:fig3}}
		\vspace{-2mm}
		\caption{Validation accuracy of LeNet-5, VGG-16, and GRU with different DL frameworks}
		\label{fig:valacc}
	\end{figure}
	
	Fig.~\ref{fig:trainacc} and \ref{fig:valacc} show the training and validation plots of LeNet-5, VGG-16, and GRU (RNN) on GPU with the same runtime configurations under different DL frameworks, respectively. We can see that all frameworks exhibit similar training behaviours, but \textsc{\small PyTorch} behaves more stably in both training and validation processes and generally has higher training accuracy compared to the other 3 frameworks in our study. 
	It is even more obvious for LeNet-5 and VGG-16, which have much larger amplitudes on these frameworks than \textsc{\small PyTorch}, as shown in Fig~\ref{fig:trainacc:fig1}, \ref{fig:trainacc:fig2} and Fig~\ref{fig:valacc:fig1}, \ref{fig:valacc:fig2}, respectively.

	\begin{table}[t]
		\centering
		\scriptsize
		\setlength\tabcolsep{1.8pt}
		\caption{Average prediction accuracy and average time costs with input data samples 10,000 on different frameworks}
		\label{tbl:prediction}
		\begin{tabular}{c|c|c|c|c|c|c|c|c} 
			\hline
			\multirow{2}{*}{\begin{tabular}[c]{@{}c@{}}{\bf DNN} \\ {\bf Models}\end{tabular}} & \multicolumn{2}{c|}{\textbf{TensorFlow} } & \multicolumn{2}{c|}{\textbf{CNTK}} & \multicolumn{2}{c|}{\textbf{PyTorch} } & \multicolumn{2}{c}{\textbf{MXNet}} \\ 
			\cline{2-9}
			& \textbf{Acc(\%)} & \textbf{Time(s)} & \textbf{Acc(\%)} & \textbf{Time(s)} & \textbf{Acc(\%)} & \textbf{Time(s)} & \textbf{Acc(\%)} & \textbf{Time(s)} \\ 
			\hline \hline
			\textbf{LeNet-1}  & 98.90 & 0.05 & 98.89 & 0.96 & 98.88 & 0.01 & 98.96 & 0.11 \\ 
			\hline
			\textbf{LeNet-5}  & 99.30 & 0.10 & 99.30 & 1.02 & 99.21 & 0.01 & 99.27 & 0.12 \\ 
			\hline
			\textbf{ResNet-20}  & 82.66 & 1.23 & 82.93 & 3.94 & 83.85 & O/M & 84.33 & 1.47 \\ 
			\hline
			\textbf{VGG-16}  & 84.70 & 3.67 & 82.77 & 11.82 & 86.12 & O/M & 86.52 & 8.86 \\ 
			\hline
			\textbf{TextCNN}  & 89.54 & 2.10 & 89.98 & 2.14 & 89.79 & 1.12 & 90.40 & 1.58 \\ 
			\hline
			\textbf{LSTM}  & 90.11 & 103.93 & 90.50 & 55.60 & 90.56 & O/M & 89.17 & 3.60 \\ 
			\hline
			\textbf{GRU}  & 90.73 & \multicolumn{1}{>{\columncolor{mygray}}c|}{\bf 85.88} & 90.92 & \multicolumn{1}{>{\columncolor{mygray}}c|}{\bf 114.69} & 91.59 & \multicolumn{1}{>{\columncolor{mygray}}c|}{\bf O/M} & 89.80 & \multicolumn{1}{>{\columncolor{mygray}}c}{\bf 3.46} \\
			\hline
		\end{tabular}
	\end{table}

	\vspace{1mm}
	\subsubsection{Prediction Accuracy}
	For each DNN model, we select one with the highest validation accuracy to conduct prediction on the testing dataset. We repeat 5 predictions on each model and find the prediction accuracy is quite similar, with time costs varying slightly. So we record the average accuracy and average time cost for evaluation.
	
	As shown in Table~\ref{tbl:prediction}, for each model, the prediction accuracy of 4 frameworks is similar with a little difference. 
	The result is reasonable because these frameworks rely on different computing libraries and provide different operator implementations (e.g., convolution operator), which finally makes the weights/biases on the same layer different from each other.
	But when it comes to time costs, models on the four frameworks behave quite differently. 
	We take GRU as an example~(marked in gray), it takes only 3.46 seconds on \textsc{\small MXNet} to predict 10,000 samples, but spends 85.88s and 114.69s on \textsc{\small TensorFlow} and \textsc{\small CNTK}, respectively. Meanwhile, it exhibits an \textit{out of memory} error under \textsc{\small PyTorch}, as marked by O/M in Table~\ref{tbl:prediction}. 
	This is mainly because \textsc{\small PyTorch} dynamically loads the data along with the graph building at each batch, without feeding in advance. Thus, \textsc{\small PyTorch} inevitably generates a large number of temporary variables in an instant, leading to the memory overflow. 
	According to the results of prediction accuracy and time costs, even given the same configuration, models under different frameworks achieve different weights/biases, resulting in different prediction accuracy and time costs. 
	This phenomenon inspires us to think if the difference is caused by the inner implementation when conduct computing.
	
	Driven by the above observations, we further investigate the prediction accuracy of different frameworks with the same weights/biases rather than the same runtime configuration.
	Specifically, we take the \textsc{\small TensorFlow} models as benchmark, and further convert them to variants fit for other frameworks, using the existing model conversion tool \textsc{\small MMdnn}~\cite{mmdnn}. 
	The outputs (i.e, the 3 variants) of \textsc{\small MMdnn} are able to share identical weights/biases with the benchmarking \textsc{\small TensorFlow} for each DNN model.
	Then we conduct predictions on them using the same testing dataset. 
	Most of the prediction accuracy across the four models are the same, but an obvious accuracy decline~(i.e, 82.66\% to 74.35\%) occurs on ResNet-20 after converted from \textsc{\small TensorFlow} to \textsc{\small CNTK}.
	
	To understand the reason, we sample the images that have inconsistent classification results by ResNet-20 on \textsc{\small TensorFlow} and \textsc{\small CNTK}.
	Taking these samples as inputs for the two models, we print the outputs of their each hidden layer.
	Strikingly, the outputs of each corresponding layer in the two models are gradually diverging as the layer deepens. 
	As shown in Table~\ref{tbl:resnet20-diff-tf-cntk:activation}, for \texttt{Activation\_1}, the first activation layer, there are only slight differences between \textsc{\small TensorFlow} and \textsc{\small CNTK} (see the pair data marked by gray).
	When it comes to the \texttt{Dense} layer (i.e., the last weight layer), the two frameworks exhibit an obvious distinction, leading to diverging classification. 
	Consider the Table~\ref{tbl:resnet20-diff-tf-cntk:dense}, the \textsc{\small TensorFlow} model predicts the image as label ``0,'' with the maximal output being ``5.7574983.''
	While the \textsc{\small CNTK} variant predicts it as label 8, with the maximal output being ``4.9673457.'' 
	Actually, we also find similar issues between other frameworks, but not obvious enough to impact the prediction logic.
	The phenomenon indicates that computation differences indeed exist between \textsc{\small TensorFlow} and \textsc{\small CNTK}, which could be amplified in models with deep layers, and introduce prediction errors.
	That should draw the attention of developers and researchers who aim to train a model on a framework and deploy on another with the help of model conversion tools.

	\begin{table}
		\scriptsize
		\centering
		\caption{The layer outputs in ResNet-20 on \textsc{\small TensorFlow} model and \textsc{\small CNTK} variant. Idx. refers to label index.}
		\label{tbl:resnet20-diff-tf-cntk}
		\subfloat[Activation\_1 Layer \label{tbl:resnet20-diff-tf-cntk:activation}]{
			\centering
			\setlength\tabcolsep{6.0pt}
			\begin{tabular}{c|c|c} 
				\hline
				\textbf{} & \textbf{TensorFlow}  & \textbf{CNTK}  \\ 
				\hline\hline
				& \multicolumn{1}{>{\columncolor{mygray}}c|}{\bf 2.50329142} & \multicolumn{1}{>{\columncolor{mygray}}c}{\bf 2.50329163}\\ 
				\cline{2-3}
				& 0.0 & 0.0 \\ 
				\cline{2-3}
				& 4.07436941 & 4.07436941 \\ 
				\cline{2-3}
				& 0.0 & 0.0 \\ 
				\cline{2-3}
				& 0.0 & 0.0 \\ 
				\cline{2-3}
				& $\cdots$  & $\cdots$  \\ 
				\cline{2-3}
				& 0.0 & 0.0 \\ 
				\cline{2-3}
				& \multicolumn{1}{>{\columncolor{mygray}}c|}{\bf 3.72458271} & \multicolumn{1}{>{\columncolor{mygray}}c}{\bf 3.72458232} \\ 
				\cline{2-3}
				& \multicolumn{1}{>{\columncolor{mygray}}c|}{\bf 0.62817883} & \multicolumn{1}{>{\columncolor{mygray}}c}{\bf 0.62817895} \\ 
				\cline{2-3}
				\multirow{-10}{*}{\rotatebox{90}{\bf Layer Output}} & 1.00697954 & 1.00697954 \\
				\hline
			\end{tabular}
		}
		\subfloat[Dense Layer~(Last Weight Layer)\label{tbl:resnet20-diff-tf-cntk:dense}]{
			\centering
			\setlength\tabcolsep{4.5pt}
			\begin{tabular}{c|c|c|c}
				\hline
				\textbf{} & \textbf{Idx.} & \textbf{TensorFlow}  & \textbf{CNTK}  \\ 
				\hline\hline
				\multirow{10}{*}{}& 0 & \multicolumn{1}{>{\columncolor{mygray}}c|}{\bf 5.7574983} & 1.9206836 \\ 
				\cline{2-4}
				& 1 & 2.6037812 & -0.5768703 \\ 
				\cline{2-4}
				& 2 & -0.6758407 & 1.2657206 \\ 
				\cline{2-4}
				& 3 &  1.3866315 & 0.73824847 \\ 
				\cline{2-4}
				& 4 & -3.348287 & -0.97601014 \\ 
				\cline{2-4}
				& 5 & -4.9494123 & -3.5435727 \\ 
				\cline{2-4}
				& 6 & -2.8659112 & -1.4083405 \\ 
				\cline{2-4}
				& 7 & -4.317035 & -2.079543 \\ 
				\cline{2-4}
				& 8 & 4.1992025 & \multicolumn{1}{>{\columncolor{mygray}}c}{\bf 4.9673457} \\ 
				\cline{2-4}
				\multirow{-10}{*}{\rotatebox{90}{\bf Layer Output}} &  9 & 2.2098625 & -0.3072039 \\
				\hline
			\end{tabular}
		}
	\end{table}
	
	\vspace{1mm}
	\noindent\fbox{
		\parbox{0.95\linewidth}{
			\textbf{Answer to RQ1}: 
			Given DL models with the same runtime configuration,
			\textsc{\small PyTorch} generally provides more stable training and validation process than \textsc{\small TensorFlow}, \textsc{\small CNTK}, and \textsc{\small MXNet} in our study. Although it is understandable that the computing differences exist across frameworks, such differences can sometimes be very obvious under certain scenarios (e.g., model conversion),
			leading to a misclassification on DL models.
			The existing model conversion between frameworks is currently not reliable due to the computing differences, which requires special attention and inspection before applying directly.
			Note that, 100\% participates in the questionnaire are interested in the quantitative differences across frameworks and the corresponding results can be used to provide development insights.
			
			\noindent \textbf{Challenge}: 
			How to identify real framework bugs according to the computing differences? How to amplify the computing differences to help find more similar issues in SE testing field?
			
			
		}
	}

	\begin{figure*}
		\centering
		\subfloat[LeNet-1]{\includegraphics[width=4.3cm]{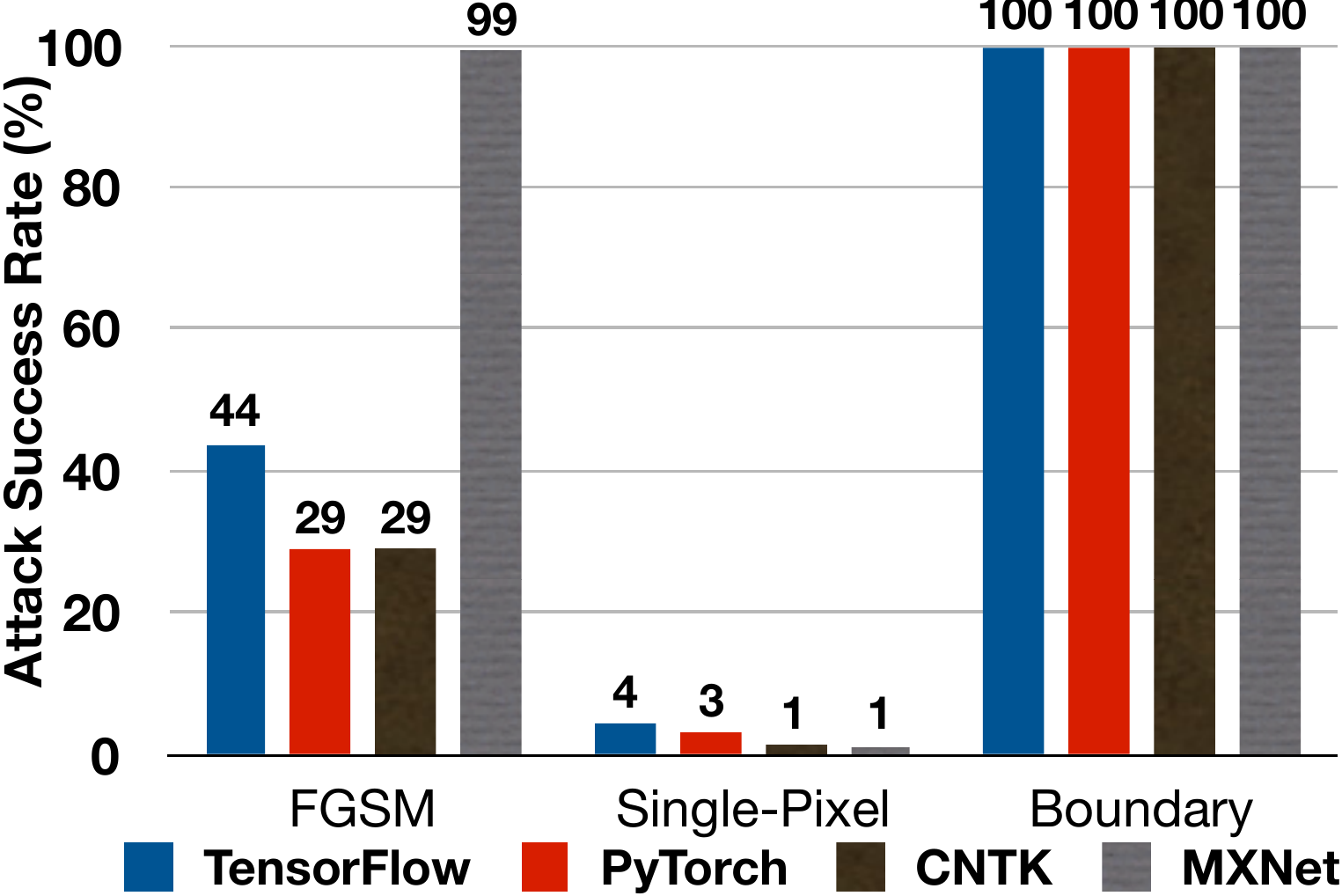}\label{fig:robustness:fig1}}
		\hspace{0.1cm}
		\subfloat[LeNet-5]{\includegraphics[width=4.3cm]{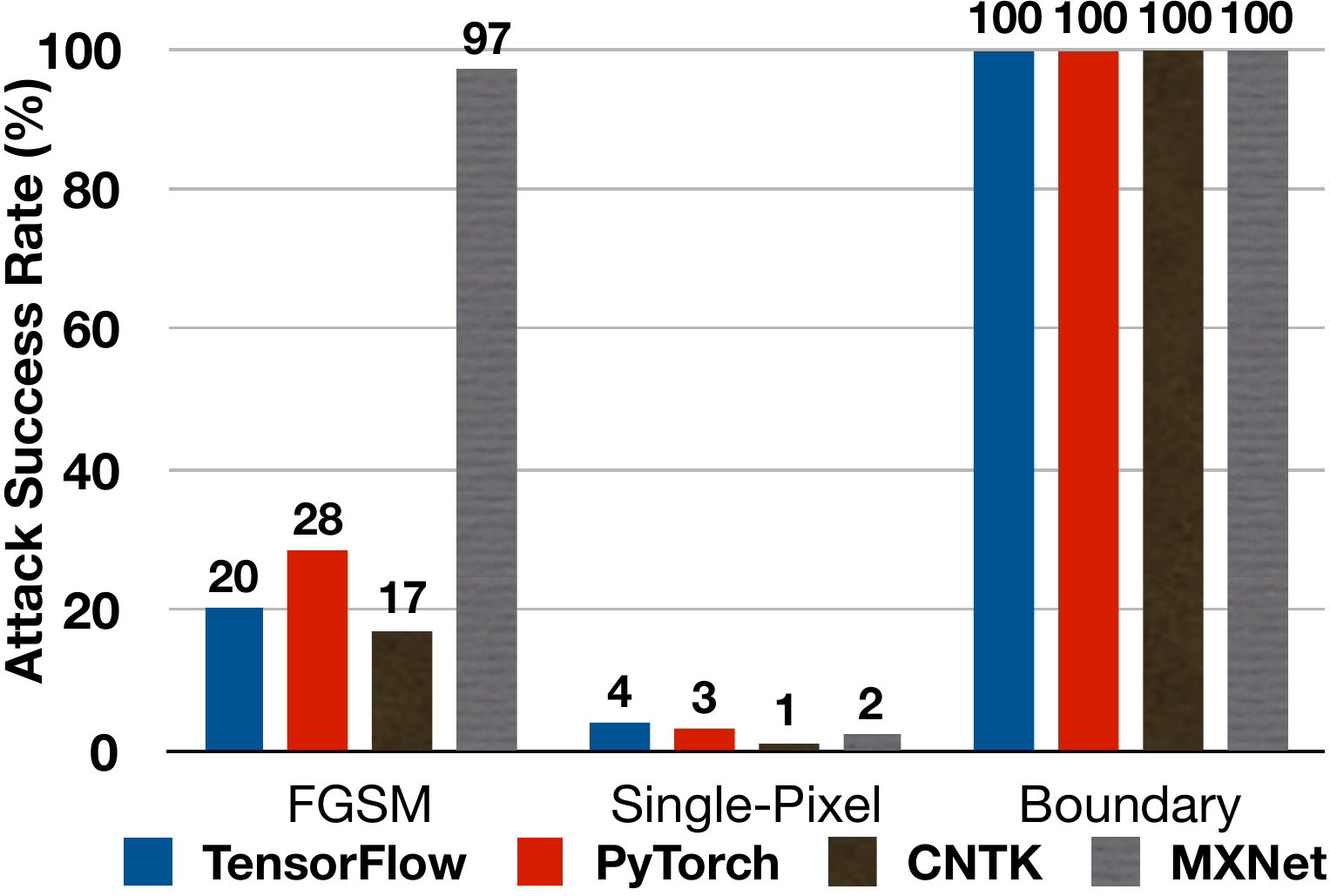}\label{fig:robustness:fig2}}
		\vspace{0.1cm}
		\subfloat[ResNet-20]{\includegraphics[width=4.3cm]{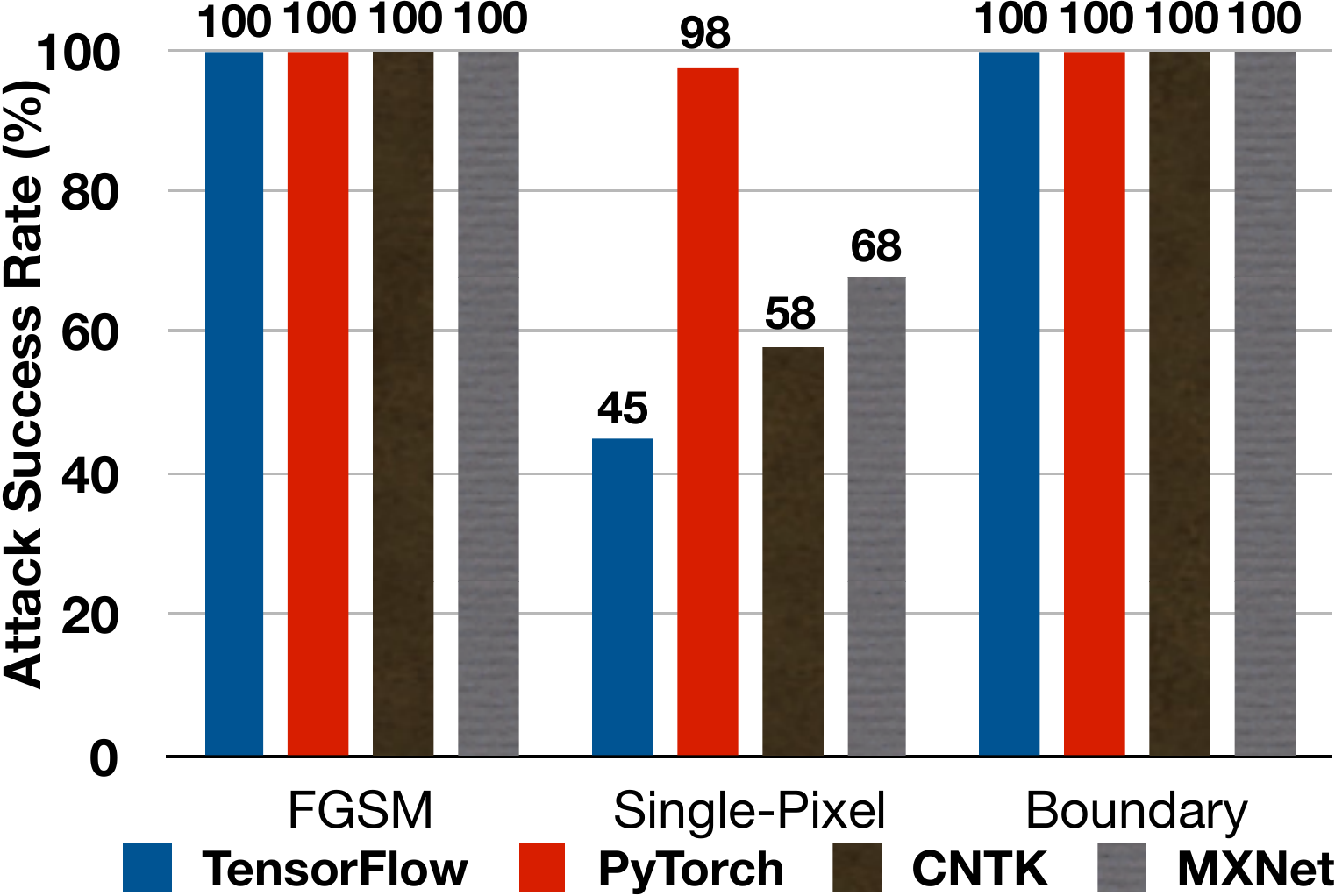}\label{fig:robustness:fig3}}
		\subfloat[VGG-16]{\includegraphics[width=4.3cm]{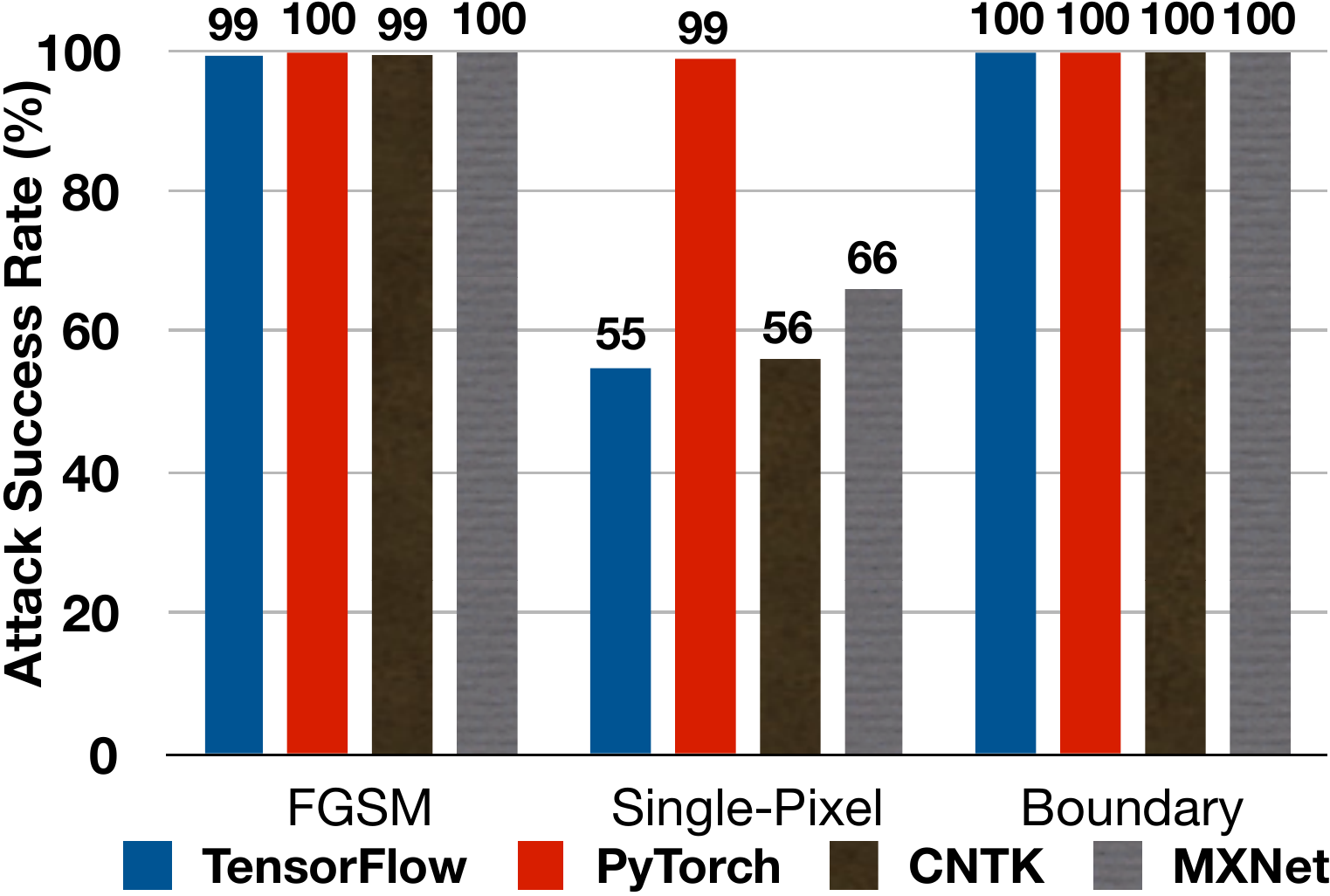}\label{fig:robustness:fig4}}
		\caption{The robustness evaluation of DL models against adversarial attacks}
		\label{fig:robustness}
	\end{figure*}

	\subsection{RQ2: Robustness of Trained Model}
	In this section, we investigate the robustness of DL models trained from different DL frameworks.
	For each model evaluated in Table~\ref{tbl:prediction},
	we examine the robustness against adversarial examples in terms of success rate, by leveraging three state-of-the-art representative adversarial attacks (i.e., FGSM~\cite{goodfellow6572explaining}, Single-Pixel-Attack~\cite{papernot2016practical}, and Boundary-Attack~\cite{brendel2017decision}). Given an input, each attack generates crafted test cases to fool the model, with following criteria:
	
	
	\begin{itemize}[noitemsep,topsep=0pt,leftmargin=*]
		\item FGSM adds perturbations along the model's gradient to craft adversarial examples.
		\item Single-Pixel-Attack adds human-unnoticeable noises on image pixel to generate adversarial
		images.  
		\item Boundary-Attack firstly performs large adversarial perturbations on input and then minimizes the $L_{2}$\text{-}norm of perturbations while staying adversarial.  
	\end{itemize}
	
	In particular, we randomly select 1,000 images in MNIST and CIFAR-10, which are correctly predicted by all the models. And these images are used as the inputs of aforementioned attacks. 
	To reduce randomness, each attack is repeated 10 times to calculate the average success rate. 
	Thus, we perform 360 configurations of attacks (4 models$\times$ 4 frameworks $\times$ 3 types of attacks $\times$ 10 repetitions).
	
	Fig.~\ref{fig:robustness} shows the average attack success rates on models
	trained from different frameworks.
	We can see Boundary-Attack achieves 100\% success rate on all DL models, because it is the most effective decision-based adversarial attack~\cite{rauber2017foolbox} to date. 
	This indicates that models trained from the state-of-the-art frameworks are still vulnerable against the advanced attacks~\cite{brendel2017decision}. 
	{Moreover, models also behave distinctly against other two attacks. Formally, we define the following equations to quantify the model robustness under attacks.}
	
	\begin{equation} 
	P(m_i,{A}) = \left\{\begin{matrix}
	\frac{S_{A}^{m_i}-min}{max-min} & min < max\\ 
	0 & min = max
	\end{matrix}\right.
	\label{equation1}
	\end{equation}
	\begin{equation} 
	R(m_i) = P(m_i,{A_1})+\ldots+P(m_i,{A_k}), \;\; k\geq 1
	\label{equation2}
	\end{equation}
	where $m_1,\ldots,m_n$ ($n\geq1$) represent the $n$ models trained from different frameworks, and ${A_k}$ are the $k$ types of attacks. $S_A^m$ represents the average success rate of attack $A$ on model $m$. Thus the {\small $min=MIN(S_A^{m_1},\ldots,S_A^{m_n})$} and {\small $max=MAX(S_{A}^{m_1},\ldots,S_{A}^{m_n})$} in Equation~\ref{equation1} indicate the minimum and maximum success rate of all models involved under attack $A$, respectively. Based on these statistics, we can compute the final robustness indicator $R(m_i)$ with Equation~\ref{equation2}, which quantifies the robustness of model $m_i$ in terms of attacks $A_1,\ldots,A_n$.
	The smaller value $R(m_i)$ is, the better robustness model $m_i$ exhibits. 
	In this study, $m_1, m_2, m_3,m_4$ represent models from \textsc{\small TensorFlow}, \textsc{\small PyTorch}, \textsc{\small CNTK}, and \textsc{\small MXNet}, respectively. And $A_1, A_2, A_3$ indicate FGSM attack, Single-Pixel attack and Boundary attack, respectively.
	
	Using above equations, we find that trained from the same runtime configurations, the \textsc{\small CNTK} models generally exhibit the best robustness compared to the models from the other three frameworks. 
	Because $R(m_3)$ comes to the minimum on LeNet-1, LeNet-5 and VGG-16, with the value being 0.01, 0.00 and 0.02, respectively. 
	{By contrast, \textsc{\small PyTorch} and \textsc{\small MXNet} are more vulnerable to attacks by adversarial samples.}
	More results can be found on our website~\cite{our}. 
	
	
	
	\vspace{1mm}
	\noindent\fbox{
		\parbox{0.95\linewidth}{
			\textbf{Answer to RQ2}:
			Given the same architecture design and runtime configuration, DL models from different frameworks exhibit diverse robustness against adversarial attacks. 
			Generally, \textsc{\small CNTK} achieves the most robust result in our evaluated settings among all the frameworks when training DL models, and models trained from \textsc{\small PyTorch} and \textsc{\small MXNet} tend to be more vulnerable to adversarial attacks.
			
			\noindent \textbf{Challenge}: {How to improve the robustness of DL models in training stage from the perspective of engineering DL frameworks?} How to develop advanced testing techniques to generate specific tests for improving robustness?
			
		}
	}
	
	\subsection{RQ3: Time and Memory Performance of Migration and Quantization on Diff-Platforms}
	In this section, we investigate the differences of capability in supporting DL software across platforms, after the model migration/quantization from the PC{/Server} platform.
	We mainly focus on the time costs and memory consumption during prediction, which are the key runtime metrics of small devices. The mobile platforms (e.g., Android OS and iOS systems) and web platforms (e.g., web browsers) are selected for evaluation. 
	
	For the mobile platform, \textsc{\small TensorFlow} and \textsc{\small Core ML} are used to migrate the trained DL models to Android and iOS platforms, respectively.
	{Specifically, for each DNN model trained by \textsc{\small TensorFlow}, we select one with the highest prediction accuracy.}
	After that, the API \texttt{TocoConverter} in \textsc{\small TensorFlow} 1.11.0 {helps migrate these trained models to the Android platforms, and the \textsc{\small TensorFlow Lite} package in Android applications provides runtime support for 
		the migrated model execution on Android devices.}
	Similarly, the \texttt{coremltools} in \textsc{\small Core ML} 2.1.0 can convert the trained models to the iOS platforms.
	
	Apart from the model migration, \textsc{\small TensorFlow} and \textsc{\small Core ML} also provide quantization techniques to optimize a model so that it can execute faster and more efficiently on mobile devices~\cite{quantization}.
	{\textsc {\small TensorFlow}} and {\textsc {\small Core ML}} provide different options to quantize the trained models for mobile platforms. 
	Since the \emph{post-training quantization} is recommended as priority by the documentation of \textsc{\small TensorFlow}\cite{post-training-quantization}, 
	{and it fixedly converts weight in trained models from 32-bits floating point to 8-bit integer using a liner weight representation.}
	We initially set the \emph{nbits} option to 8 and select the \emph{linear mode} in \textsc{\small Core ML} for all DL models to ensure the consistency. 
	Additionally, since the VGG-16 model cannot be quantized to 8-bits in practice \cite{numpy-error}, {we only use a 16-bits quantization for VGG-16.}
	In this study, 6 representative real mobile devices (i.e., HUAWEI Mate 20 X, HUAWEI Nexus 6P, and Motorola Nexus 6 with Android OS and iPhone 8, iPhone 6S, and iPad Pro with iOS) are selected for evaluation.
	
	For the web platform, \textsc{\small TensorFlow.js} 0.14.2 is used to migrate the trained \textsc{\small TensorFlow} models to the format which could be loaded by web browsers.
	The web platform refers to the browsers on PC, rather than on mobile devices. We select 3 mainstream browsers (i.e., Chrome, Firefox, and Safari) for web evaluation, and run them on a Macbook Pro. 
	\begin{table}
		\centering
		\scriptsize
		\setlength\tabcolsep{4.4pt}
		\caption{Prediction accuracy and time cost on different mobile devices}
		\label{tbl:mobile}
		
		\begin{tabular}{c|c|c|c|c|c|c|c}
			\hline
			\multirow{3}{*}{\begin{tabular}[c]{@{}c@{}}{\bf DNN} \\ {\bf Mod.}\end{tabular}} & 
			\multirow{3}{*}{\textbf{Plat.}} & 
			\multirow{3}{*}{\textbf{Device}} & 
			\multirow{3}{*}{{\bf Quan.}} & \multirow{3}{*}{\begin{tabular}[c]{@{}c@{}}{\bf Size}\end{tabular}} & \multicolumn{2}{c|}{\textbf{Original}} 
			& \textbf{Generated}  \\ \cline{6-8} 
			&  &  &  &  & 
			\begin{tabular}[c]{@{}c@{}}{\bf Acc.} \\ {(\%)}\end{tabular} & 
			\begin{tabular}[c]{@{}c@{}}{\bf Pred.} \\ {\bf Time}(s)\end{tabular} & \begin{tabular}[c]{@{}c@{}}{\bf Acc.} \\ {(\%)}\end{tabular} \\ \hline
			\hline
			\multirow{13}{*}{\rotatebox{90}{\textbf{LeNet-1}}} & PC & Server & No & 16KB & 98.70 & 0.05 & 87.42   \\ \cline{2-8} 
			& \multirow{6}{*}{\rotatebox{90} {Android}} & \multirow{2}{*}{Nexus 6} & No & 15KB & 98.70 & 5.33 & 87.42 \\ \cline{4-8} 
			&  &  & Yes & 5.4KB & \multicolumn{1}{>{\columncolor{mygray}}c|}{\bf 98.69} & 3.80 & \multicolumn{1}{>{\columncolor{mygray}}c}{\bf 82.32} \\ \cline{3-8} 
			&  & \multirow{2}{*}{Nexus 6P} & No & 15KB & 98.70 & 4.19 & 87.42 \\ \cline{4-8} 
			&  &  & Yes & 5.4KB & \multicolumn{1}{>{\columncolor{mygray}}c|}{\bf 98.69} & 3.32 & \multicolumn{1}{>{\columncolor{mygray}}c}{\bf 82.32}\\ \cline{3-8} 
			&  & \multirow{2}{*}{Mate 20X} & No & 15KB & 98.70 & 2.09 & 87.42 \\ \cline{4-8} 
			&  &  & Yes & 5.4KB  & \multicolumn{1}{>{\columncolor{mygray}}c|}{\bf 98.69} & 1.51 & \multicolumn{1}{>{\columncolor{mygray}}c}{\bf82.32} \\ \cline{2-8} 
			& \multirow{6}{*}{iOS} & \multirow{2}{*}{iPhone 6S} & No & 14KB & 98.70 & 235.66 & 86.51 \\ \cline{4-8} 
			&  &  & Yes & 4.5KB & 98.70 & \multicolumn{1}{>{\columncolor{mygray}}c|}{\bf 238.27} & \multicolumn{1}{>{\columncolor{mygray}}c}{\bf 81.46} \\ \cline{3-8} 
			&  & \multirow{2}{*}{iPhone 8} & No & 14KB & 98.70 & 121.78 & 86.54 \\ \cline{4-8} 
			&  &  & Yes & 4.5KB & \multicolumn{1}{>{\columncolor{mygray}}c|}{\bf 98.65} & \multicolumn{1}{>{\columncolor{mygray}}c|}{\bf 123.56} & \multicolumn{1}{>{\columncolor{mygray}}c}{\bf 81.49} \\ \cline{3-8} 
			&  & \multirow{2}{*}{iPad Pro} & No & 14KB & 98.70 & 145.92 & 86.51 \\ \cline{4-8} 
			&  &  & Yes & 4.5KB & \multicolumn{1}{>{\columncolor{mygray}}c|}{\bf 98.66} & \multicolumn{1}{>{\columncolor{mygray}}c|}{\bf 147.41} & \multicolumn{1}{>{\columncolor{mygray}}c}{\bf 81.46}  \\ \hline
			\multirow{13}{*}{\rotatebox{90}{\bf LeNet-5}} & PC & Server & No & 178KB & 99.13 & 0.10 & 89.24 \\ \cline{2-8} 
			& \multirow{6}{*}{\rotatebox{90}{Android}} & \multirow{2}{*}{Nexus 6} & No & 176KB & 99.13 & 8.31 & 89.24 \\ \cline{4-8} 
			&  &  & Yes & 50KB & 99.13 & 5.30 & \multicolumn{1}{>{\columncolor{mygray}}c}{\bf 83.31} \\ \cline{3-8} 
			&  & \multirow{2}{*}{Nexus 6P} & No & 176KB & 99.13 & 6.16 & 89.24 \\ \cline{4-8} 
			&  &  & Yes & 50KB & 99.13 & 4.26 & \multicolumn{1}{>{\columncolor{mygray}}c}{\bf 83.31} \\ \cline{3-8}
			&  & \multirow{2}{*}{Mate 20X} & No & 176KB & 99.13 & 5.28 & 89.24 \\ \cline{4-8} 
			&  &  & Yes & 50KB & 99.13 & 1.17 & \multicolumn{1}{>{\columncolor{mygray}}c}{\bf 83.31} \\ \cline{2-8}
			& \multirow{4}{*}{iOS} & \multirow{2}{*}{iPhone 6S} & No & 175KB & 99.13 & 245.62 & 88.87 \\ \cline{4-8} 
			&  &  & Yes & 47KB & 99.13 & \multicolumn{1}{>{\columncolor{mygray}}c|}{\bf 248.92} & \multicolumn{1}{>{\columncolor{mygray}}c}{\bf 82.96} \\ \cline{3-8} 
			&  & \multirow{2}{*}{iPhone 8} & No & 175KB & 99.13 & 128.84 & 88.96 \\ \cline{4-8} 
			&  &  & Yes & 47KB & \multicolumn{1}{>{\columncolor{mygray}}c|}{\bf 99.09} & \multicolumn{1}{>{\columncolor{mygray}}c|}{\bf 130.47} & \multicolumn{1}{>{\columncolor{mygray}}c}{\bf 83.04} \\ \cline{3-8}
			&  & \multirow{2}{*}{iPad Pro} & No & 175KB & 99.13 & 153.47 & 88.87 \\ \cline{4-8} 
			&  &  & Yes & 47KB & \multicolumn{1}{>{\columncolor{mygray}}c|}{\bf 99.09} & \multicolumn{1}{>{\columncolor{mygray}}c|}{\bf 153.70} & \multicolumn{1}{>{\columncolor{mygray}}c}{\bf 81.61} \\ \hline
			\multirow{13}{*}{\rotatebox{90}{\bf ResNet-20}} & PC & Server & No & 1.1MB & 83.05 & 1.23 & 77.70 \\ \cline{2-8} 
			& \multirow{6}{*}{\rotatebox{90}{Android}} & \multirow{2}{*}{Nexus 6} & No & 1.1MB & 83.05 & 565.30 & 77.70 \\ \cline{4-8} 
			&  &  & Yes & 290KB & 83.06 & 320.41 & \multicolumn{1}{>{\columncolor{mygray}}c}{\bf 73.49} \\ \cline{3-8} 
			&  & \multirow{2}{*}{Nexus 6P} & No & 1.1MB & 83.05 & 495.21 & 77.70 \\ \cline{4-8} 
			&  &  & Yes & 290KB & 83.06 & 262.24 & \multicolumn{1}{>{\columncolor{mygray}}c}{\bf 73.49}\\ \cline{3-8}
			&  & \multirow{2}{*}{Mate 20X} & No & 1.1MB & 83.05 & 240.67 & 77.70 \\ \cline{4-8} 
			&  &  & Yes & 290KB & \multicolumn{1}{>{\columncolor{mygray}}c|}{\bf 82.93} & 113.05 & \multicolumn{1}{>{\columncolor{mygray}}c}{\bf 73.49} \\ \cline{2-8}
			& \multirow{6}{*}{iOS} & \multirow{2}{*}{iPhone 6S} & No & 1.1MB & 83.09 & 374.73 & 76.28 \\ \cline{4-8} 
			&  &  & Yes & 281KB & 83.05 & \multicolumn{1}{>{\columncolor{mygray}}c|}{\bf 383.49} & \multicolumn{1}{>{\columncolor{mygray}}c}{\bf 72.15} \\ \cline{3-8} 
			&  & \multirow{2}{*}{iPhone 8} & No & 1.1MB & {\bf 83.04} & 224.23 & 77.03 \\ \cline{4-8} 
			&  &  & Yes & 281KB & 83.02 & \multicolumn{1}{>{\columncolor{mygray}}c|}{\bf 229.41} & \multicolumn{1}{>{\columncolor{mygray}}c}{\bf 72.86} \\ \cline{3-8}
			&  & \multirow{2}{*}{iPad Pro} & No & 1.1MB & 83.08 & 230.35 & 76.26 \\ \cline{4-8} 
			&  &  & Yes & 281KB & 83.06 & 
			\multicolumn{1}{>{\columncolor{mygray}}c|}{\bf 232.78} & \multicolumn{1}{>{\columncolor{mygray}}c}{\bf 72.13}  \\ \hline
			\multirow{13}{*}{\rotatebox{90}{\bf VGG-16}} & PC & Server & No & 129MB & 84.20 & 3.67 & 79.25  \\ \cline{2-8} 
			& \multirow{6}{*}{\rotatebox{90}{Android}} & \multirow{2}{*}{Nexus 6} & No & 129MB & 84.20 & \emph{\textbf{2432.51}} & 79.25 \\ \cline{4-8} 
			&  &  & Yes & 33MB & \multicolumn{1}{>{\columncolor{mygray}}c|}{\bf 84.19} & \emph{\textbf{823.15}} &   \multicolumn{1}{>{\columncolor{mygray}}c}{\bf 75.28} \\ \cline{3-8} 
			&  & \multirow{2}{*}{Nexus 6P} & No & 129MB & 84.20 & \emph{\textbf{2909.95}} & 79.25 \\ \cline{4-8} 
			&  &  & Yes & 33MB & \multicolumn{1}{>{\columncolor{mygray}}c|}{\bf 84.19} & \emph{\textbf{1996.54}} & \multicolumn{1}{>{\columncolor{mygray}}c}{\bf 75.28} \\ \cline{3-8}
			&  & \multirow{2}{*}{Mate 20X} & No & 129MB & 84.20 & 1595.82 & 79.25 \\ \cline{4-8} 
			&  &  & Yes & 33MB & \multicolumn{1}{>{\columncolor{mygray}}c|}{\bf 84.19} & 322.60 & \multicolumn{1}{>{\columncolor{mygray}}c}{\bf 75.28} \\ \cline{2-8}
			& \multirow{6}{*}{iOS} & \multirow{2}{*}{iPhone 6S} & No & 129MB & {\bf 84.19} & 1699.90 & {77.54} \\ \cline{4-8} 
			&  &  & Yes & 65MB & 84.22 & \multicolumn{1}{>{\columncolor{mygray}}c|}{\bf 1768.87} & \emph{\textbf{77.56}} \\ \cline{3-8} 
			&  & \multirow{2}{*}{iPhone 8} & No & 129MB & 84.21 & 1143.95 & {79.05} \\ \cline{4-8} 
			&  &  & Yes & 65MB & 84.21 & \multicolumn{1}{>{\columncolor{mygray}}c|}{\bf 1210.45} & \emph{\textbf{78.93}} \\ \cline{3-8}
			&  & \multirow{2}{*}{iPad Pro} & No & 129MB & {\bf 84.19} & 939.63 & {77.55} \\ \cline{4-8} 
			&  &  & Yes & 65MB & 84.22 & \multicolumn{1}{>{\columncolor{mygray}}c|}{\bf 964.00} & \emph{\textbf{77.57}} \\ \hline
		\end{tabular}
		\begin{tablenotes}
			\item DNN Mod.: DNN models;  Plat.: platform;  Quan.: quantization;  Acc: accuracy; Pred. Time: prediction time; Original: original dataset; Generated: generated dataset
		\end{tablenotes}
	\end{table}

	\begin{table}
		\centering
		\scriptsize
		\setlength\tabcolsep{2.6pt}
		\caption{Prediction performance of DNN models on MNIST and CIFAR-10 with different web browsers}
		\label{tbl:browser1}
		\begin{tabular}{c|c|c|c|c|c|c|c|c|c}
			\hline
			&  &  &  & \multicolumn{3}{c|}{\textbf{Original Data}} & \multicolumn{3}{c}{\textbf{Generated Data}} \\ \cline{5-10} 
			\multirow{-2}{*}{\begin{tabular}[c]{@{}c@{}}{\bf DNN} \\ {\bf Mod.}\end{tabular}} & \multirow{-2}{*}{\textbf{Plat.}} & \multirow{-2}{*}{\begin{tabular}[c]{@{}c@{}}{\bf Size}\end{tabular}} & \multirow{-2}{*}{\textbf{Browser}} & \begin{tabular}[c]{@{}c@{}}{\bf Acc.} \\ {(\%)}\end{tabular} & \begin{tabular}[c]{@{}c@{}}{\bf Pred.} \\ {\bf Time}\end{tabular} & \begin{tabular}[c]{@{}c@{}}{\bf System} \\ {\bf Memory}\end{tabular} & \begin{tabular}[c]{@{}c@{}}{\bf Acc.} \\ {(\%)}\end{tabular} & \begin{tabular}[c]{@{}c@{}}{\bf Pred.} \\ {\bf Time}\end{tabular} & \begin{tabular}[c]{@{}c@{}}{\bf System} \\ {\bf Memory}\end{tabular} \\ \hline \hline
			& PC & 52KB & - & 98.90 & 0.05 & - & 79.37 & 0.09 & - \\ \cline{2-10} 
			&  &  & Chrome & 98.90 & 0.68 & - & 79.37 & 2.14 & - \\ \cline{4-10} 
			&  &  & Firefox & 98.90 & 1.32 & - & 79.37 & 2.68 & - \\ \cline{4-10} 
			\multirow{-4}{*}{\rotatebox{90}{\textbf{LeNet-1}}} & \multirow{-3}{*}{Web} & \multirow{-3}{*}{20KB} & Safari & 98.90 & 0.99 & - & 79.37 & 2.92 & - \\ \hline
			& PC & 380KB & - & 99.30 & 0.10 & - & 78.60 & 0.16 & - \\ \cline{2-10} 
			&  &  & Chrome & 99.30 & 0.93 & - & 78.60 & 2.59 & - \\ \cline{4-10} 
			&  &  & Firefox & 99.30 & 1.72 & - & 78.60 & 3.15 & - \\ \cline{4-10} 
			\multirow{-4}{*}{\rotatebox{90}{\textbf{LeNet-5}}} & \multirow{-3}{*}{Web} & \multirow{-3}{*}{184KB} & Safari & 99.30 & 1.44 & - & 78.60 & 3.52 & - \\ \hline
			& PC & 2.4MB & - & 82.66 & 1.23 & - & 68.97 & 1.85 & - \\ \cline{2-10} 
			&  &  & Chrome & \multicolumn{1}{>{\columncolor{mygray}}c|}{\bf 77.08} & 22.80 & 2.41GB & \multicolumn{1}{>{\columncolor{mygray}}c|}{\bf 61.96} & 31.07 & 2.46GB \\ \cline{4-10} 
			&  &  & Firefox & \multicolumn{1}{>{\columncolor{mygray}}c|}{\bf 77.08} & 25.22 & 3.52GB & \multicolumn{1}{>{\columncolor{mygray}}c|}{\bf 61.96} & 42.41 & - \\ \cline{4-10} 
			\multirow{-4}{*}{\rotatebox{90}{\textbf{ResNet-20}}} & \multirow{-3}{*}{Web} & \multirow{-3}{*}{1.1MB} & Safari & \multicolumn{1}{>{\columncolor{mygray}}c|}{\bf 77.08} & 79.92 & \multicolumn{1}{>{\columncolor{mygray}}c|}{\bf 4.37GB} &\multicolumn{1}{>{\columncolor{mygray}}c|}{\bf 61.96} & 81.72 & \multicolumn{1}{>{\columncolor{mygray}}c}{\bf 6.49GB} \\ \hline
			& PC & 258MB & - & 84.70 & 3.67 & - & 67.60 & 3.95 & \multicolumn{1}{>{\columncolor{mygray}}c}{\bf $*$} \\ \cline{2-10} 
			&  &  & Chrome & 84.70 & 139.83 & 2.06GB & 67.60 & 167.50 & 2.52GB \\ \cline{4-10} 
			&  &  & Firefox & 84.70 & 153.08 & 3.30GB & 67.60 & 300.85 & \multicolumn{1}{>{\columncolor{mygray}}c}{\bf $*$} \\ \cline{4-10} 
			\multirow{-4}{*}{\rotatebox{90}{\textbf{VGG-16}}} & \multirow{-3}{*}{Web} & \multirow{-3}{*}{129MB} & Safari & 84.70 & 156.74 & \multicolumn{1}{>{\columncolor{mygray}}c|}{\bf 4.66GB} & 67.60 & 490.46 & \multicolumn{1}{>{\columncolor{mygray}}c}{\bf 8.69GB} \\ \hline
		\end{tabular}
		\begin{tablenotes}
			\item Mod.: models; Plat.: platform; Size: model size; Acc: accuracy; Pred. Time: prediction time(s); Mem.: Memory
			\item $*$ means the exception on Firefox due to ``allocation size overflow.''
		\end{tablenotes}
	\end{table}

	Table~\ref{tbl:mobile}, \ref{tbl:browser1}, and \ref{tbl:browser2} show the results of prediction accuracy and time cost on different platforms and the effects of migration and quantization for mobile devices and web browsers. 
	For mobile platforms, four CNN models are evaluated, because we cannot convert the RNN models (i.e., LSTM and GRU) to mobile platforms due to the ``unsupported operation'' error~\cite{bug}, which indicates that the current supporting of DL tasks on mobile platforms is unfledged.
	Note that quantization is only performed on mobile devices in our study,
	{because there is no quantization support for web platforms until now.} 
	For web browsers, all the seven trained DL models are selected to migrate. 
	We record the \textit{System Memory} consumption in prediction process. 
	Notably, we do not record the system memory consumption and energy of mobile devices since the record process is inaccurate due to many limitations such as the impacts of mobile system and runtime environment.
	
	\subsubsection{Time Performance} 
	
	For mobile platform, Android and iOS devices exhibit different time performance which depends on DL model type. As shown in Table~\ref{tbl:mobile} (Column \textit{Pred. Time}), for the LeNet-1 and LeNet-5, there is a big difference in time performance on iOS and Android devices. 
	Android devices take less than 9s to predict while iOS devices spend more than 100s, and even up to 248.92s (i.e., iPhone 6S for LeNet-5). Different from the LeNet family, iOS devices predict faster than Android devices for ResNet-20 and VGG-16. It seems that as the complexity of the model increases, the performance advantage of iOS devices gradually emerges.
	
	In terms of the prediction time of quantized models, predicting on Android devices after quantization is faster than the original model, the improvement is more obvious for complex models (e.g., ResNet-20 and VGG-16). 
	Strikingly, quantization on iOS slows down the prediction speed a little as shown in Column \textit{Original-Pred. Time} (in gray) in Table~\ref{tbl:mobile}, which is an overall trend and confused phenomenon. 
	Note that, we have reported the issue to \textsc{\small Core ML}.
	
	As shown in Table~\ref{tbl:mobile}, we use two types of mobile devices (i.e., Nexus 6 and Nexus 6P) to observe the time performance. Most cases reflect the trend (i.e., Nexus 6P is an upgraded version of Nexus 6, therefore, the prediction time on Nexus 6P should be less than Nexus 6.).
	However, as shown in Column \textit{Original-Pred. Time} (in bold italic), Nexus 6P spends more time than Nexus 6 when running VGG-16, which indicates the platforms' capability of supporting DL software is likely related to specific model type.
	
	For the time on web browsers, Chrome generally outperforms the other two  browsers in our study. As shown in Column \textit{Original Data-Pred. Time} in Table~\ref{tbl:browser1} and \ref{tbl:browser2}, it spends less time on Chrome than Firefox and Safari in predicting the same amount of testing data. There is only one anomaly occurs for VGG-16, which Chrome costs 284.62s longer than the 191.45s on Safari. 
	
	\begin{table}[t]
		\centering
		\scriptsize
		\setlength\tabcolsep{4.8pt}
		\caption{Prediction performance of DNN models on IMDb with different web browsers.}
		\label{tbl:browser2}
		\begin{tabular}{c|c|c|c|c|c|c}
			\hline
			\multirow{3}{*}{\begin{tabular}[c]{@{}c@{}}{\bf DNN} \\ {\bf Models}\end{tabular}} & \multirow{3}{*}{\textbf{Platform}} & \multirow{3}{*}{\begin{tabular}[c]{@{}c@{}}{\bf Model} \\ {\bf Size}\end{tabular}} & \multirow{3}{*}{\textbf{Browser}} & \multicolumn{3}{c}{\textbf{Original Data}} \\ \cline{5-7} 
			&  &  &  & \begin{tabular}[c]{@{}c@{}}{\bf Accuracy} \\ {(\%)}\end{tabular}  & \begin{tabular}[c]{@{}c@{}}{\bf Pred.} \\ {\bf Time (s)}\end{tabular} & \begin{tabular}[c]{@{}c@{}}{\bf System} \\ {\bf Memory}\end{tabular} \\ \hline \hline
			& PC & 40MB & - & 89.54 & 2.10 & - \\ \cline{2-7} 
			&  &  & Chrome & 89.54 &  65.57 & 253.65MB \\ \cline{4-7} 
			&  &  & Firefox & 89.54 & 67.52 & 417MB \\ \cline{4-7}
			\multirow{-4}{*}{\rotatebox{90}{\textbf{TextCNN}}} & \multirow{-3}{*}{Web} & \multirow{-3}{*}{13MB} & Safari & 89.54 & 69.33 & \multicolumn{1}{>{\columncolor{mygray}}c}{\bf 1.07GB} \\ \hline 
			& PC & 48MB & - & 90.11 & 103.93 & - \\ \cline{2-7} 
			&  &  & Chrome & 90.11 & 248.37 & 210.2MB \\ \cline{4-7} 
			&  &  & Firefox & 90.11 & 375.20 & 1.24GB \\ \cline{4-7} 
			\multirow{-4}{*}{\rotatebox{90}{\textbf{LSTM}}} & \multirow{-3}{*}{Web} & \multirow{-3}{*}{16MB} & Safari & 90.11 & 260.49 & \multicolumn{1}{>{\columncolor{mygray}}c}{\bf 1.83GB} \\ \hline 
			& PC & 45MB & - & 90.73 & 85.88 & - \\ \cline{2-7} 
			&  &  & Chrome & 90.73 & 284.62 & 232.9MB \\ \cline{4-7} 
			&  &  & Firefox & 90.73 & 471.81 & 1.37GB \\ \cline{4-7} 
			\multirow{-4}{*}{\rotatebox{90}{\textbf{GRU}}} & \multirow{-3}{*}{Web} & \multirow{-3}{*}{15MB} & Safari & 90.73 & 191.45 & \multicolumn{1}{>{\columncolor{mygray}}c}{\bf 1.64GB} \\ \hline
		\end{tabular}
	\end{table}

	\subsubsection{Memory Performance}
	As shown in Table~\ref{tbl:browser1} and \ref{tbl:browser2}, apart from prediction time, we also record the system memory consumption on web platforms.
	System Memory consumption is a more representative metric than prediction time, when evaluating the supporting capability for DL software. 
	Note that we do not record the system memory on LeNet-1 and LeNet-5,
	{because their fleeting prediction processes make it hard to record the corresponding system memory}.
	
	As shown in Column \textit{System Memory}, predicting on web browsers are memory-consuming for all models. 
	Among the 3 browsers, Safari consumes the largest system memory. And according to our observation, the huge consumption of system memory has affected the performance of the host computer. 
	Although the memory performance of Firefox and Chrome is better than Safari, their memory overheads still reach several GB size in most cases. 
	For example, the memory overhead is over 2.4GB when running ResNet-20 on the 3 browsers, indicating the browsers' capability of supporting DL software is not satisfactory till now. 
	Combined the metrics of prediction time and system memory consumption,
	{Chrome exhibits the best performance in supporting DL tasks, which could be a better choice when running DL applications on browsers.} 
	
	
	\vspace{2mm}
	\noindent\fbox{
		\parbox{0.93\linewidth}{
			\textbf{Answer to RQ3}: 
			Different platform devices hold different time and memory performance in capability of supporting DL software. 
			For mobile devices, Android devices take much less time than iOS devices for simple DNN models. However, as the complexity of the model increases, iOS devices achieve better time performance.
			Moreover, the capability of supporting DL software on mobile platform is likely related to the types of specific DNN models.
			For web platforms, Chrome generally outperforms others in both prediction time cost and system memory consumption in our study. The overall performance for web DL software is unsatisfactory, especially running complex DL models.
			
			\noindent  \textbf{Challenge}: {How to reduce the time cost memory consumption after model migration and quantization? 
				How to further test the performance of different platforms when deploying and running DL software systematically?}

		}
	}

	\begin{table}
		\centering
		\scriptsize
		\caption{The layer outputs of ResNet-20 on PC and Chrome. Idx. refers to label index.}
		\vspace{3mm}
		\label{tbl:resnet20-diff}
		
		\subfloat[Conv2D Layer \label{tbl:resnet20-diff:conv2d}]{
			\centering
			\setlength\tabcolsep{6.0pt}
			\begin{tabular}{c|c|c} 
				\hline
				& \textbf{PC}  & \textbf{Chrome}  \\ 
				\hline\hline
				\multirow{10}{*}{} & 1.78371441 & 1.78371441 \\ 
				\cline{2-3}
				& \multicolumn{1}{>{\columncolor{mygray}}c|}{\bf -1.47859037} & \multicolumn{1}{>{\columncolor{mygray}}c}{\bf -1.47859049} \\ 
				\cline{2-3}
				& \multicolumn{1}{>{\columncolor{mygray}}c|}{\bf 0.57163376} & \multicolumn{1}{>{\columncolor{mygray}}c}{\bf 0.57163370} \\ 
				\cline{2-3}
				& \multicolumn{1}{>{\columncolor{mygray}}c|}{\bf -0.01394593} & \multicolumn{1}{>{\columncolor{mygray}}c}{\bf -0.01394595} \\ 
				\cline{2-3}
				& -2.42872572 & -2.42872572 \\ 
				\cline{2-3}
				& $\cdots$ & $\cdots$ \\ 
				\cline{2-3}
				& \multicolumn{1}{>{\columncolor{mygray}}c|}{\bf 0.003037585} & \multicolumn{1}{>{\columncolor{mygray}}c}{\bf 0.003037592} \\ 
				\cline{2-3}
				& \multicolumn{1}{>{\columncolor{mygray}}c|}{\bf -0.43665951} & \multicolumn{1}{>{\columncolor{mygray}}c}{\bf -0.43665954} \\ 
				\cline{2-3}
				& 0.08801108 & 0.08801108 \\ 
				\cline{2-3}
				\multirow{-10}{*}{\rotatebox{90}{\bf Layer Output}} & \multicolumn{1}{>{\columncolor{mygray}}c|}{\bf 0.174682378} & \multicolumn{1}{>{\columncolor{mygray}}c}{\bf 0.174682394} \\
				\hline
			\end{tabular}
		}
		\subfloat[Dense Layer~(Last Weight Layer)\label{tbl:resnet20-diff:dense}]{
			\centering
			\setlength\tabcolsep{4.5pt}
			\begin{tabular}{c|c|c|c}
				\hline
				& \textbf{Idx.} & \textbf{PC}  & \textbf{Chrome}  \\ 
				\hline\hline
				\multirow{10}{*}{}& 0 &  3.94917989 & -1.03813171 \\
				\cline{2-4}
				& 1 & -5.77517033 & -3.22286654 \\
				\cline{2-4}
				& 2 & 5.10022831 & \multicolumn{1}{>{\columncolor{mygray}}c}{\bf 3.76064563} \\
				\cline{2-4}
				& 3 &  -2.74950528 & 1.56391966 \\ 
				\cline{2-4}
				& 4 & 3.04771161 & -1.47325206 \\ 
				\cline{2-4}
				& 5 & -2.04092622 & 0.52656615 \\
				\cline{2-4}
				& 6 & -6.07451582 & -0.17675309  \\ 
				\cline{2-4}
				& 7 & \multicolumn{1}{>{\columncolor{mygray}}c|}{\bf 8.46383476} & 2.48092437 \\ 
				\cline{2-4}
				& 8 & -0.23961751 & 1.38548911 \\ 
				\cline{2-4}
				\multirow{-10}{*}{\rotatebox{90}{\bf Layer Output}} & 9 & -3.68060803 & -3.80600476 \\
				\hline
			\end{tabular}
		}
	\end{table}
	
	\vspace{2mm}
	\subsection{RQ4: Accuracy of Migration and Quantization on Diff-Platforms}
	In this section, we investigate the prediction accuracy after DL model migration and quantization on different platforms (i.e., mobile and web platforms).
	
	\subsubsection{Model Migration for Different Platforms}\label{sec:rq4-1}
	As shown in Table~\ref{tbl:mobile}, \ref{tbl:browser1} and \ref{tbl:browser2}, for each DNN model, we first compare the accuracy of each model without quantization on different mobile platforms (marked as \textit{No} in Column \textit{Quan.}).
	We find that the DNN model size does not change a lot after migrating the \textsc{\small TensorFlow} model to mobile platforms. 
	However, the size of each model for web platform decreases by a large margin.
	
	Using the original test data, the accuracy of the mobile migration is almost unchanged, the biggest change comes from the data of iPhone 6S on VGG-16 (i.e., 84.19 vs. 84.20) and iPhone 8 on ResNet-20 (i.e., 83.04 vs. 83.05). 
	Similarly, the accuracy of web migration generally shares the same trend. However, a significant accuracy decline occurs on all 3 browsers for ResNet-20 (i.e., 77.08 vs. 82.66), as shown in Table~\ref{tbl:browser1}.
	To analyze and explain the reason for this severe compatibility issue, we first compare the model structure and weights between the two platforms (i.e., PC and web) and confirm that they share the same properties of them. So we further inspect the outputs of each layer for ResNet-20 on PC and web browsers.
	Strikingly, given the same input image, we find the outputs of each layer on PC and web browsers are different. Moreover, the deeper the layer is, the more obvious difference they exhibit. 
	
	We take Chrome as example to give an in-depth comparative analysis on a certain image. As shown in Table~\ref{tbl:resnet20-diff:conv2d},
	for \texttt{Conv2D}, the first weight layer connecting to the input, there are only slight differences between Chrome and PC (see the pair data marked by gray).
	When it comes to the \texttt{Dense} layer (i.e., the last weight layer), the two platforms exhibit an obvious distinction, leading to a misclassification on Chrome.
	As shown in Table~\ref{tbl:resnet20-diff:dense}, the PC model predicts the image as label ``7,'' with the maximal output being ``8.46383476.''
	While Chrome predicts it as label 2, with the maximal output being ``3.76064563.'' Other two browsers also show the similar behaviours.
	The result indicates that browsers differ from PC in inner-model computing, leading to the accuracy decline on ResNet-20.
	Actually, similar compatibility issues also occur on LeNet-1, LeNet-5, and VGG-16 when migrated from PC to browsers, although the final prediction logic are not been influenced.
	We reported these issues to the team of \textsc{\small TensorFlow.js}, and the developers have acknowledged as a real bug {when WebGL handles \(1\times1\) \texttt{Conv2D} kernels,} and will fix it in the new release version.
	

	\vspace{3mm}
	\noindent\fbox{
		\parbox{0.93\linewidth}{
			\textbf{Answer to RQ4-1}: 
			The prediction accuracy on original data has not been affected much by the migration process. However, compatibility issues persist in model migration from PC to browsers (e.g., 77.08 vs. 82.66 on ResNet-20). 
			Even worse, there still exists a obvious
			difference on computation mechanism between PC and web browsers, leading to a computing distinction of each layer within the model, {which has been acknowledged and confirmed by the team of \textsc{\small TensorFlow.js}}. This result explains why the industry has failed to meet expectations after model migration based on our online questionnaire, which provides a reasonable explanation for the industrial developers. 
		}
	}
	
	\vspace{3mm}
	\subsubsection{Model Quantization for Mobile Platforms}\label{sec:rq4-2}
	Considering the models marked as \textit{Yes} in Column \textit{Quan.} in Table~\ref{tbl:mobile}, the model size decreases roughly 50\% to 75\% after quantization. 
	It saves much storage and memory for mobile devices, exactly according with the intentions for designing quantization. 
	The quantization process does not significantly affect the prediction accuracy on original testing data. Specifically, the biggest change comes from HUAWEI Mate 20 X on ResNet-20 (i.e., 82.93 vs. 83.05). Even in some cases, the accuracy of quantized model is higher. For example, the accuracy of the quantized ResNet-20 model on other Android devices increases by 0.01\% and the quantized VGG-16 model on iPhone 6S and iPad Pro rises by 0.03\%.
	
	
	\vspace{2mm}
	\noindent\fbox{
		\parbox{0.93\linewidth}{
			\textbf{Answer to RQ4-2}: 
			Quantization does not affect the prediction accuracy obviously.
			{Prediction on Android devices after quantization is faster than the original model, and the improvement is more significant for complex models. {Strikingly, quantization on iOS devices slows down the prediction speed, {which deserves further optimization for \textsc{\small Core ML}}.}}
		}
	}
	
	\vspace{3mm}
	\subsubsection{Migration and Quantization on Generated Data}
	According to section \ref{sec:rq4-1} and \ref{sec:rq4-2}, the migration/quantization does not affect the prediction accuracy obviously, there still exist some cases that the accuracy decreases, especially for the quantization process. 
	The results of accuracy in above two sections are based on the original testing data. 
	To further investigate the \emph{quality} of migrated/quantized models, 
	we combine the existing tools \textsc{\small TensorFuzz}~\cite{odena2018tensorfuzz} and \textsc{\small DeepHunter}~\cite{xie2019deephunter} as data generator. 
	We generate a large-scale testing data by using MNIST and CIFAR-10 as inputs to capture the differential behaviors between the PC model and the migrated/quantized model. 
	25,000 mutated MNIST data are created for LeNet-1 and LeNet-5, respectively. 
	28,000 mutated CIFAR-10 data are generated for ResNet-20 and VGG-16, respectively. 
	We generate 106,000 samples for both mobile and browser in total.
	
	
	We run the migrated models repeatedly on our generated data for the two platforms.
	As shown in Table~\ref{tbl:mobile}, the prediction accuracy of migrated models remain unaltered on Android devices, consistent to the result on original testing data.
	However, iOS devices go through a relatively obvious accuracy decline on our generated testing data.
	For example, iPhone 6S, iPhone 8 and iPad Pro achieve 76.28\%, 77.03\% and 76.26\% accuracy on ResNet-20 respectively, which are less than the 77.70\% on server. 
	In addition, LeNet-1 and LeNet-5 show the similar phenomenon, which 
	indicates the migration process on iOS devices suffers from reliability issues on the generated data.
	As for web platforms, the accuracy of ResNet-20 still drops more than 5\% accuracy (i.e., {61.96\% vs. 68.97\%}), which agrees with the result on the original data (i.e., 77.08\% vs. 82.66\%). The similar result on generated data validates our findings about the compatibility issues in migration process. 
	
	Strikingly, as shown in Column \textit{Generated-Acc.} (in gray), the accuracy of all quantized models has a significant decline, indicating the reliability of a quantizated model is unsatisfactory to date. However, the different results on the two datasets (i.e., original testing data and generated testing data) show that it is hard to trigger the reliability issue with the original widely-used datasets.
	Last but not least, for iOS devices, the accuracy of quantized models on VGG-16 only drops a little, since we follows a different modes (i.e., 32-bits to 16-bits), compared to other three models when reducing the floating point.
	To investigate whether the accuracy of quantized models is relevant to the value of $nbits$ in float reduction, we further obverse the ResNet-20 as an example, and configure the $nbits$ as 8 and 4.
	{Results show that the accuracy gradually declines with a decreasing bit value.}
	The accuracy are {77.57\%, 74.42\% and 8.53\%} corresponding to the floating point from 32-bits to 16-bits, 8-bits and 4-bits on iPad Pro, respectively.
	
	
	
	\vspace{2mm}
	\noindent\fbox{
		\parbox{0.93\linewidth}{
			\noindent{\textbf{Remarks for inspection of generated data}}: 
			(1) The accuracy of migrated models does not change in our evaluation
			on Android devices, while has a relatively obvious decline on iOS devices. As for the web platforms, the results (i.e., compatibility bugs) are consistent to that on original data.
			(2) The accuracy of all quantized models has a significant decline on our generated testing data, which indicates the quantization process still suffers from severe {reliability} issues tested by generated data. Meanwhile, the decline is correlated with the value $nbits$ when reducing the floating point on iOS devices.
			{(3) 
				Furthermore, we conduct statistical analysis~\cite{wilcoxon} on the accuracy-dropping cases in Column \textit{Generated} after quantization of Table~\ref{tbl:mobile}. The results give a $p<0.05$,
				indicating there exists a statistically significant difference in accuracy on generated data, which reconfirms the reliability issues.}
			
			\noindent{\textbf{Challenge}}: {How to detect and fix the compatibility issues/bugs when migrating the trained models to web platforms and iOS devices, and the reliability issues when quantizing the trained models to mobile platforms?
			}
			
		}
	}

	\subsection{Threats to Validity}
	(1) The DNN models and datasets we used might not be complete, thus our findings are not general for all situations. But we select models with CNN/RNN architecture from various domains, ranging from image classification to textual sentiment analysis. Moreover, the datasets contain diverse types, including gray, color images and textual review, to reduce such a threat.
	(2) The selected versions of DL frameworks in our study might not be complete. However, we do not focus on the multi-version evolution, but on revealing challenges/issues that developers and researchers need to consider in development and deployment processes.
	(3) Three Android devices and three iOS devices with fixed versions are used to study the prediction performance on mobile platforms. 
	We mainly focus on the performance change after the model migration/quantization from PC to mobile devices, the impacts of mobile hardware and mobile system version on prediction performance are beyond the scope of this work.
	
	\vspace{2mm}
	\section{Related Work}
	In this section, we review the related work in two aspects: study of deep learning frameworks and platforms. 
	Actually, for the studies of model migration and quantization on different deep learning platforms (i.e., mobile devices and browsers), to the best our knowledge, we take the first step towards this research field.
	Several deep learning benchmarking studies have been done on the basic results of deep learning frameworks~\cite{bahrampour2015comparative,awan2017depth,coleman2017dawnbench,shatnawi2018comparative} such as the influence of different hardwares and training accuracy and time, and also compared different frameworks using their default configuration settings and parameters~\cite{liu2018benchmarking}. However, there lacks a systemic study on the different impacts that various deep learning frameworks under the same runtime configuration or same model weights/biases have on the deep learning software development and deployment, and also lacks an investigation on quantitative showing the differences of frameworks for developers and researchers.
	
	\subsection{Study of DL Platforms}
	Kaoru et al.~\cite{ota2017deep} made a survey on deep learning for mobile multimedia and introduced the low-complexity deep learning algorithms, an optimized software framework for mobile environments and the specialized hardware for supporting the computationally expensive processes of deep network training and inference. 
	AI-Benchmark~\cite{aibenchmark} proposed a AI performance ranking for current mainstream mobile phones. Nine testing tasks such as object recognition and face recognition are used as criteria for performance comparison.
	Alsing et al.~\cite{alsing2018mobile} summarized the latest mobile object detection methods using \textsc{TensorFlow Lite} and analyzed the performance and latency payoff of different deep learning models on mobile devices.
	Wang et al.~\cite{wang2018deep} provided an overview of the current achievements about mobile deep learning technologies and applications. 
	Xu et al.~\cite{xu2019dlsmartphone} conducted an empirical study on a large-scale Android apps to investigate how deep learning technique is adopted in practice.
	{Ma et al.~\cite{ma2019moving} investigated seven JavaScript-based deep learning frameworks and measured their performance gaps when running different deep learning tasks on Chrome. 
		However, we focus on the difference of supporting capabilities when deep learning tasks are deployed on various web browsers (i.e., Chrome, Firefox, and Safari).}
	
	\balance
	\subsection{Study of DL Frameworks}
	The rapid emergence of deep learning frameworks attracts researchers' attention on the performance of deep learning frameworks. 
	The most related work is from Liu et al.~\cite{liu2018benchmarking}, they conducted a comparative study of three frameworks (i.e., \textsc{\small TensorFlow}, \textsc{\small Caffe}, and \textsc{\small Torch}). However, they observed from various aspects such as the impacts of default settings and dataset-dependent default settings, and framework-dependent default settings in deep learning frameworks, which are totally different from us.
	Moreover, Bahrampour et al.~\cite{bahrampour2016comparative} presented a comparative study on four deep learning frameworks (i.e., \textsc{Caffe}, \textsc{Neon}, \textsc{Theano}, and \textsc{Torch}). They evaluated these frameworks from three aspects (i.e., extensibility, hardware utilization, and speed).
	Shams et al.~\cite{shams2017evaluation} analyzed \textsc{Caffe}, \textsc{TensorFlow} and \textsc{Apache SINGA} over several hardware environments. In order to investigate the performance, they measured the time per training iteration and the number of images trained with in a millisecond for comparison.
	Kochura et al.~\cite{kochura2017comparative} compared the basic features (i.e., GPU
	support, GUI, operating systems, and language support) of \textsc{TensorFlow}, \textsc{Deep Learning4j} and \textsc{H2O} and conducted throughout performance tests. In particular, \textsc{H20} was tested under single threaded mode and multi-threaded mode.
	Li et al.~\cite{li2016evaluating} evaluated the energy efficiency of CNNs on CPUs and GPUs by calculating the energy and power consumption of ten deep learning frameworks (\textsc{K20-Torch}, \textsc{Tx-Caffe}, etc.).
	Shaohuai et al.~\cite{shi2016benchmarking} calculated the time per mini-batch with different threads (i.e., 1, 2, 4, 8) and deep neural network models (\textsc{FCN-S}, \textsc{\small ResNet-50}, etc.) within \textsc{Caffe}, \textsc{CNTK}, \textsc{TensorFlow}, \textsc{MXNet} and \textsc{Torch}. 
	Amershi et al.~\cite{microsoft-study} provided
	a description of how several Microsoft software engineering teams work on developing AI applications.
	Apart from the above work on deep learning frameworks, 
	{several work focused on the bug detection of deep learning frameworks}. For example,  Zhang et al.~\cite{zhang2018empirical} studied 175 \textsc{TensorFlow} bugs and examied the root causes of these bugs. Pham el al.~\cite{phamcradle} proposed \textsc{CRADLE}, a new approach that cross-checks multiple backends to find and localize bugs in deep learning software libraries. 
	
	\vspace{1mm}
	\subsection{Deep Learning Testing}
	Some existing techniques have been proposed to detect the problems/issues during deep learning development and deployment. DeepXplore~\cite{pei2017deepxplore} and DeepGauge~\cite{ma2018deepgauge} proposed the new testing criteria for deep learning testing. DeepTest~\cite{tian2018deeptest}, DeepHunter~\cite{xie2019deephunter} and TensorFuzz~\cite{odena2018tensorfuzz} proposed coverage-guided testing techniques, which mainly focus on feedforward neural networks. DeepStellar~\cite{du2019deepstellar} is proposed to perform the quantitative analysis for recurrent neural networks (RNN). DeepMutation~\cite{ma2018deepmutation} adopts the mutation testing techniques to evaluate the quality of test data for a deep neural network. In addition, DiffChaser~\cite{diffchaser} proposed a differential testing technique to capture the minor disagreements of two deep neural networks. The approach can be applied to detect the issues of deep neural networks caused by deep learning platforms and frameworks.
	
	{In summary, compared to these studies on deep learning frameworks and platforms,} our study conducted a systematic study including training performance and prediction accuracy when given the same runtime configuration or model weights/biases, adversarial robustness, model migration and quantization on different frameworks and platforms, and the capabilities and reliability of supporting deep learning software on different platforms.
	Moreover, we not only conduct evaluations on the PC/Server platform, but also shift the testing on the real mobile devices and web browsers. 
	Meanwhile, based on our study, we also reported several real deep learning software bugs and provide useful guidance for deep learning developers and researchers. In addition, our study motivates many new research directions such as deep learning software bug detection when model migrated and quantized under different deep learning platforms and model conversion.

	\vspace{1mm}
	\section{Conclusion}
	In this paper, we initiate the first step to investigate how existing deep learning frameworks and platforms influence the development and deployment of deep learning software.
	Our study provides many practical guidelines for developers and researchers under different scenarios for different research communities.
	Given the same model weights/biases, an obvious accuracy decline occurs when the model is converted from one framework to another.
	The compatibility and reliability issues and accuracy loss would arise when migrating and quantizing a deep learning model from the PC platform to other platforms,
	and the accuracy loss is due to several deep learning software bugs we found.
	In addition, the universal deep learning solutions across platforms are desperately on demand, especially for mobile and web platforms. 
	This study makes the first step along this direction towards building universal deep learning software across various platforms based on our practical guidelines.
	We hope our work draws the attention of deep learning software community, altogether to address the urgent demands towards the new challenges in deep learning software development and deployment processes.
	
	\vspace{1mm}
	\section{Acknowledgments}
	This research was partially been supported by the National Science Foundation of China (No. 61872262, 61572349).
	It was also sponsored by the National Research Foundation, Prime Ministers Office, Singapore under its National Cybersecurity R\&D Program (Award No. NRF2018NCR-NCR005-0001),
	National Satellite of Excellence in Trustworthy Software System (Award No. NRF2018NCR-NSOE003-0001) administered by the National Cybersecurity R\&D Directorate, and JSPS KAKENHI Grant 19K24348, 19H04086, and Qdai-jump Research Program NO.01277.
	\clearpage
	\balance
	
	\bibliographystyle{IEEEtran}
	\bibliography{ref}
	
\end{document}